\documentclass{article}

\PassOptionsToPackage{numbers, compress}{natbib}

\usepackage[eandd, final]{neurips_2026}

\usepackage[utf8]{inputenc} 
\usepackage[T1]{fontenc}    
\usepackage{hyperref}       
\usepackage{url}            
\usepackage{booktabs}       
\usepackage{amsfonts}       
\usepackage{nicefrac}       
\usepackage{microtype}      
\usepackage{multirow}
\usepackage{amsmath}
\usepackage{graphicx}
\usepackage{tabularx}      
\usepackage{array}         
 \usepackage{enumitem}
\newcommand{\ans}[1]{\textbf{\textcolor{ansok}{#1}}}
\usepackage{caption}
\usepackage{subcaption}
\usepackage{comment}

\usepackage{amssymb}
\usepackage{makecell}

\usepackage[ruled,vlined,linesnumbered]{algorithm2e}
\usepackage[table]{xcolor}
\newcommand{\Path}[1]{{\color{pathblue}\textit{Path.}~#1}}
\newcommand{\Qq}[1]{\textit{Q.}~#1}
\usepackage[most]{tcolorbox}
\usepackage{placeins}

\definecolor{ansok}{RGB}{0,128,0}   
\definecolor{pathblue}{RGB}{0, 70, 180}   
\definecolor{relA}{RGB}{31,119,180}    
\definecolor{relB}{RGB}{214, 39, 40}   
\definecolor{relC}{RGB}{44, 160, 44}   

\usepackage{xcolor}\usepackage{xspace}\usepackage{listings}
\usepackage[most]{tcolorbox}
\definecolor{onecclr}{RGB}{30,90,170}
\definecolor{sbdclr}{RGB}{180,90,30}
\newcommand{\OC}[1]{\textcolor{onecclr}{\textbf{#1}}}
\newcommand{\SD}[1]{\textcolor{sbdclr}{\textbf{#1}}}
\lstdefinestyle{promptstyle}{basicstyle=\ttfamily\footnotesize,
  breaklines=true,breakatwhitespace=true,columns=fullflexible,
  keepspaces=true,showstringspaces=false,mathescape=false,
  escapeinside={(*}{*)}}
\tcbset{promptbox/.style={colback=gray!4,colframe=gray!50!black,
  boxrule=0.4pt,arc=2pt,breakable,enhanced jigsaw,
  fonttitle=\bfseries\small,left=4pt,right=4pt,top=4pt,bottom=4pt}}

\title{HiddenPathQA: A Benchmark for Knowledge Graph Question Answering When Questions Hide Their Paths}

\author{%
  Sumin Jo, Edward Choi \\
  KAIST \\
  \texttt{\{ekrxjwh2009,edwardchoi\}@kaist.ac.kr}
}

\newcommand{\methodname}{\textbf{Ours}\xspace}
\usepackage{xspace}
\newcommand{\sbd}{\textsc{TieredHolistic}\xspace}
\newcommand{\onec}{\textsc{FlatHolistic}\xspace}
\newcommand{\bench}{\textsc{HiddenPathQA}\xspace}

\begin{document}

\maketitle


\begin{abstract}
Knowledge graph question answering (KGQA) systems are typically evaluated on questions whose surface tokens lexically expose the gold KG path. Users, however, may ask questions without knowing the underlying schema: “Which condition could Andrew have due to family history?” compresses a 4-hop chain “Andrew $\xrightarrow{\text{father}}$ parent $\xrightarrow{\text{father}}$ grandparent $\xrightarrow{\text{medical condition}}$ disease $\xrightarrow{\text{subclass of}}$ hereditary disorder” into a single phrase (family history) whose tokens name none of those relations. We show that this collapse of path–question alignment is a blind spot of existing KGQA evaluation. Three major Large Language Model (LLM)–KG paradigms—hop-by-hop KG exploration, plan-then-retrieve, and similarity-based retrieval—all implicitly rely on the question’s surface leaking the gold path, an assumption rarely tested in practice.
To study this failure mode, we make three contributions. (i) We introduce \textbf{HiddenPathQA}, a manually verified benchmark of 1,055 multi-hop questions spanning 2–6 hops and multiple domains, constructed so that the question surface does not reveal the gold path’s relations. (ii) We propose \textbf{Holistic Path Selection}, which compares the question against candidate relation chains in the local subgraph as complete units, replacing irreversible per-hop relation scoring with globally informed chain-level reasoning. We develop two implementations, TieredHolistic and FlatHolistic. (iii) We conduct a pseudoword stress test that replaces entity names with meaningless placeholder tokens, allowing us to distinguish KG-grounded reasoning from entity-level memorization. Holistic Path Selection outperforms six strong KGQA baselines(ToG, PoG, FiDeLiS, R2-KG,
KAPING, KARPA) by 8.8–17.9 accuracy points across two LLM backbones and both the original and pseudoword settings. Its advantage under pseudoword replacement shows that the gains cannot be explained by entity-level memorization alone. The dataset and code are publicly available at \href{https://github.com/ekrxjwh2009/HiddenPathQA}{\texttt{GitHub}} and \href{https://huggingface.co/datasets/suminJo/HiddenPathQA}{\texttt{HuggingFace}}
\end{abstract}

\section{Introduction}
\subsection{Three paradigms, one shared assumption}
\label{sec:intro:paradigms}
Recent prompt-based KG-augmented LLM methods for KGQA can be 
characterized along three broad paradigms.\footnote{We focus on 
prompt-based (training-free or lightly-tuned) methods that operate over a 
given KG at inference time. Methods that require KG-specific fine-tuning 
(e.g., GNN-based retrievers) or that frame KGQA as semantic parsing 
to executable logical forms lie outside the scope of this 
characterization.}
\textbf{(i) Hop-by-hop KG exploration} approaches, which form the 
dominant paradigm in recent LLM-based KGQA research, such as 
ToG~\citep{tog}, PoG~\citep{pog}, FiDeLiS~\citep{fidelis}, and 
R2-KG~\citep{r2kg}, let the LLM interact with the KG over multiple 
hops, at each hop selecting a small set of promising relations and 
expanding the search frontier.
\textbf{(ii) Plan-then-retrieve} approaches, such as KARPA~\citep{karpa} 
and RoG~\citep{rog}, first prompt the LLM to produce a relation-level 
plan or candidate relation paths and then ground the plan to executable 
KG paths via similarity between the LLM's plan tokens and KG 
relation labels.
\textbf{(iii) Similarity-based evidence retrieval} approaches, such as
KAPING~\citep{kaping} and DiFaR~\citep{difar}, encode the question and
KG facts (triples or paths) into a shared dense embedding space and
retrieve the top-ranked evidence by vector similarity, which is then
provided to the LLM as context.
Although these paradigms differ algorithmically, they share an implicit 
dependence on the surface form of the question: retrieval or 
exploration is easier when the question contains lexical cues that 
either name the relevant relations, guide relation-path planning, or 
place the question close to the supporting KG facts in embedding 
space. We refer to this shared property as the \emph{path--question 
lexical alignment assumption}.

This assumption is largely satisfied by existing benchmarks. Consider
the MetaQA~\citep{metaqa} query \emph{``the films
\textcolor{red!75!black}{written by} the
\textcolor{blue!75!black}{screenwriter} of [To Catch a Thief] were
\textcolor{green!50!black}{directed by} who?''}, which all but spells
out the relation sequence
\emph{\textcolor{blue!75!black}{written by}} $\to$
\emph{\textcolor{red!75!black}{written by}} $\to$
\emph{\textcolor{green!50!black}{directed by}}. Under such lexical
alignment, all three paradigms work well: stepwise pruning recovers
the correct relation at each hop from the question surface alone,
plan-then-retrieve produces a relation plan that grounds to the gold
sequence, and embedding retrieval easily ranks the matching triples on
top.

\subsection{When the assumption breaks: a real-world example}
\label{sec:intro:hidden}

In practice, users do not know the KG schema, and their questions are
rarely as KG-aligned as benchmark queries suggest. Consider the
real-world query
\emph{``Which condition could Andrew have due to
\textcolor{relA}{family history}?''}, whose actual KG path is
\begin{quote}
Andrew $\to$ (\emph{\textcolor{relA}{father}}) $\to$ Andrew's parent
$\to$ (\emph{\textcolor{relA}{father}}) $\to$ Andrew's grandparent
$\to$ (\emph{\textcolor{relA}{medical condition}}) $\to$ Disease $\to$
(\emph{\textcolor{relA}{subclass of}}) $\to$ hereditary disorder.
\end{quote}
No surface token in the question corresponds to \emph{father},
\emph{medical condition}, or \emph{subclass of}: the single phrase
``family history'' \textbf{compresses an entire 4-hop chain into one
expression}. We call this regime \emph{alignment collapse}. Under it, each paradigm is pulled toward a different wrong attractor.
The 1-hop relations of \emph{Andrew} include \emph{father} (gold)
alongside \emph{occupation}, \emph{given name}, \emph{family name},
\emph{family}, \emph{cause of death}, and \emph{medical
condition}...etc. (\textbf{i}) \emph{hop-by-hop KG exploration} prompts the LLM to prune
this set against ``family history'' at each hop independently. The
LLM may retain \emph{father} at hop~1 via semantic association, but
at hop~2---now expanding from \emph{Andrew's parent}---the question
offers no fresh signal to repeat \emph{father}; having already moved
``through family,'' the LLM transitions to medically salient
relations such as \emph{medical condition} or \emph{cause of death}.
The chain breaks not because hop~1 is hard, but because per-hop
scoring has no mechanism to recognize that the same relation(\emph{father}) must be
applied twice---a structural blind spot independent of the LLM's
strength. (\textbf{ii}) \emph{Plan-then-retrieve} paraphrases ``family history''
as a medical concept and emits a hop-1 plan such as \emph{cause of
death} or \emph{medical condition}; embedding-based grounding then
locks the plan onto the wrong relation.
(\textbf{iii}) \emph{Similarity-based retrieval} suffers from
token-level overlap between ``family history'' and the spurious
relations \emph{family name} and \emph{family}, which dominate the
top-$k$ and push the gold first-hop triple out.

The three paradigms thus fail through different attractors---LLM
commitment failure, semantic paraphrase, and lexical-overlap
distractor---but the cause is the same: the question carries no signal
pointing to \emph{parents}, leaving each paradigm to substitute its
own prior. Without prior knowledge of the KG schema, the only reliable way to
recover the gold path is to consider the entire local graph around
\emph{Andrew} jointly, rather than commit early on the basis of any
local fragment of it. We verify these mechanisms quantitatively
across paradigms in §\ref{res:failureofbaseline}.

\subsection{Our benchmark: \bench}
\label{sec:intro:bench}

\begin{figure*}[t]
    \centering
    \includegraphics[width=\textwidth]{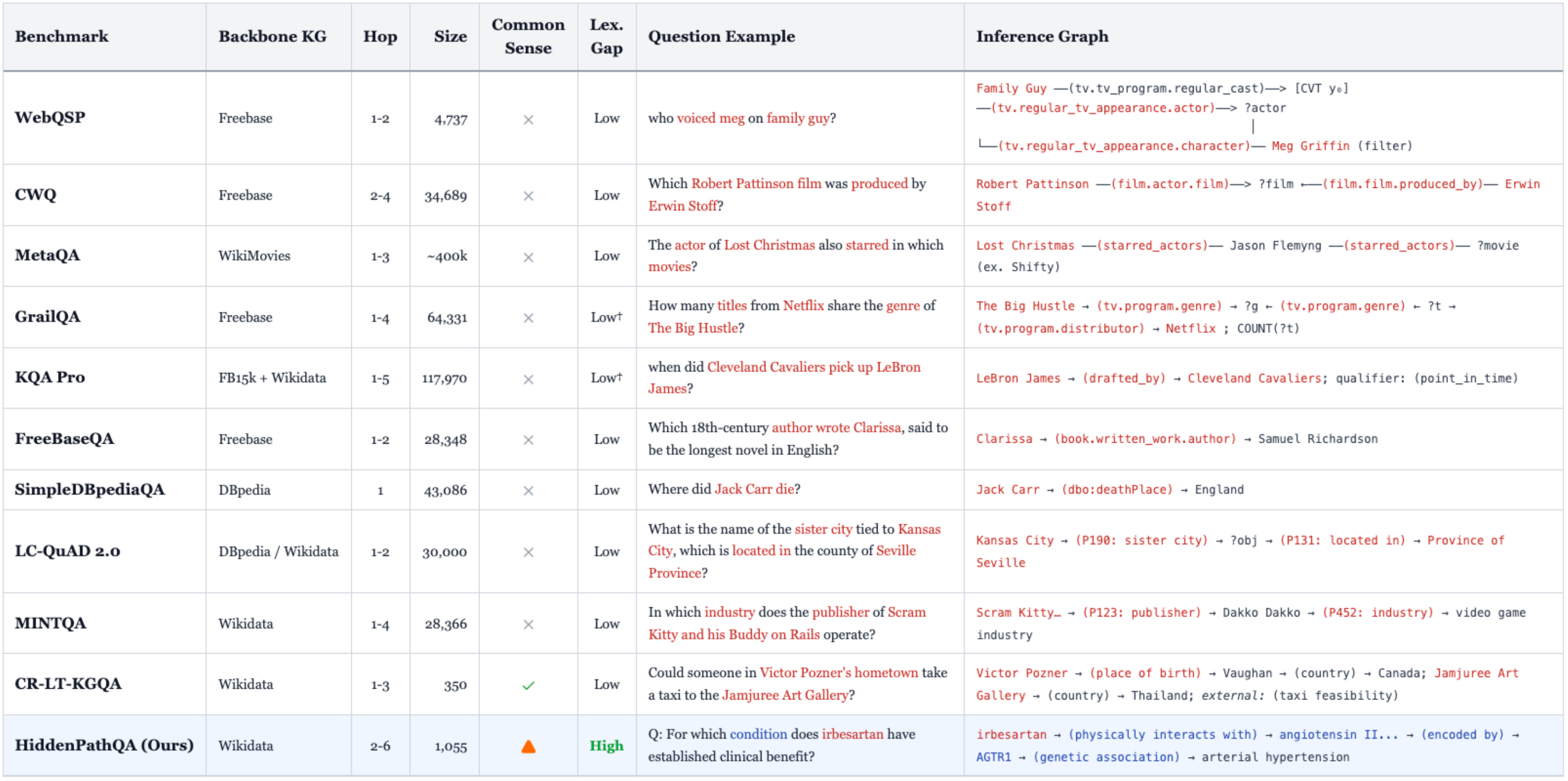}
    \caption{Overview of KGQA benchmarks alongside \bench.
\textcolor{red}{\textbf{Red}} = entities/relations appearing
(almost) verbatim in both the question and the inference graph
(\emph{low} lexical gap). \textcolor{blue}{\textbf{Blue}} =
entities/relations appearing in only one of the two
(\emph{high} lexical gap). \emph{Low}$^{\dagger}$: questions
derived from a logical form, where surface overlap is reduced
but path--question alignment is preserved at the logical-form
level. \textcolor{orange}{$\blacktriangle$}: \bench does not
test commonsense, but uses compressed expressions (e.g.,
\emph{aunt-in-law}) that the LLM must interpret to recover
the correct relation chain.}
    \label{fig:benchmark_compare}
\end{figure*}

What remains genuinely unsolved is whether KGQA systems can recover the correct path when the question hides it—
when multiple hops are compressed into a single
phrase, when the relevant relations share no surface form with the
question, and when the LLM cannot fall back on lexical shortcuts.
Existing KGQA benchmarks, despite covering a wide range of difficulty
axes, do not explicitly control how much of the gold KG path is leaked
through the question's surface tokens; we provide a detailed comparison
in \S\ref{sec:related} and Figure~\ref{fig:benchmark_compare}. We introduce \bench, a human-validated benchmark of 1{,}055 multi-hop(2--6 hops)
KGQA questions whose design deliberately
\emph{restricts path--question lexical alignment}: the relations on the
gold path are not present as tokens in the question. This forces models
to reason over the full chain rather than match keywords. We instantiate
the principle through three question types---\emph{compressed-only},
\emph{cause-and-effect}, and \emph{intermediate-node} (defined and
exemplified in \S\ref{sec:bench:types})---and present each item as a
five-way multiple-choice question.

\subsection{Our method: Holistic Path Selection}
\label{sec:intro:method}

We address this failure mode head-on. Instead of committing to
relations on the basis of \emph{local} cues, we present \emph{all} candidate relation chains around
the topic entity to the LLM at once and let it select the answer
chain \emph{globally} in a small number of inference steps.  We call this paradigm \textbf{Holistic Path Selection} and provide two implementations \sbd and \onec, differing only in
how chains are exposed across depths
(\S\ref{sec:method:pipeline}). Crucially, neither variant prunes
candidates at intermediate hops: every chain reachable up to the
current depth remains visible to the LLM in one view, so no relation
is irreversibly discarded based on hop-local cues. The simplicity is
intentional: when the question surface is compressed, hop-level
local cues are noise, and global chain semantics is the actual
signal.

\section{Related Benchmarks}
\label{sec:related}

\paragraph{Axes of difficulty in prior benchmarks.}
Figure~\ref{fig:benchmark_compare} compares recent KGQA benchmarks
alongside \bench. \emph{Hop depth and multi-hop composition:}
 WebQSP~\citep{webqsp} pioneered semantic parsing over Freebase~\citep{freebase} with
short multi-hop queries; CWQ~\citep{cwq} extends WebQSP with
compositional questions of higher reasoning depth; MetaQA~\citep{metaqa}
templated multi-hop questions in a closed movie-domain KG.
\emph{Compositional and schema generalization:}
GrailQA~\citep{grailqa} explicitly probes i.i.d., compositional, and
zero-shot generalization across Freebase schema items.
\emph{Logical and SPARQL-style operations:}
KQA~Pro~\citep{kqapro} targets multi-step logical reasoning with
aggregation and comparison; LC-QuAD~2.0~\citep{lcquad2} pairs natural
questions with executable SPARQL queries over DBpedia~\citep{dbpedia} and Wikidata~\citep{wikidata}.
\emph{Large-scale factoid coverage:}
FreeBaseQA~\citep{freebaseqa} and SimpleDBpediaQA~\citep{simpledbpedia}
supply large pools of shallow factoid questions for broad-coverage
evaluation. \emph{Knowledge unfamiliar to LLMs:} MINTQA~\citep{mintqa}
isolates new and long-tail knowledge---entities that appear rarely
in pretraining corpora and are thus unlikely to be memorized by
pretrained LLMs. \emph{Commonsense$+$KG fusion:}
CR-LT-KGQA~\citep{crltkgqa} requires combining KG facts with
commonsense reasoning over long-tail entities.

\paragraph{The missing axis: path--question alignment.}
Across these axes, questions tend to retain lexical traces of the
gold KG path on the surface, as illustrated in
Figure~\ref{fig:benchmark_compare}: relation labels and intermediate
entity names recur as tokens in the question, so a model can recover
much of the path without traversing the KG. This leakage is built
into how the questions are constructed, not a side effect: standard
pipelines---whether they start from a logical form, a sampled KG
path, or a human-written question later mapped to the KG---never
restrict which relation names or intermediate entities are allowed
to appear in the final question. No prior benchmark adds such a
restriction, so none can measure how KGQA methods perform when this
surface alignment is removed.
This restriction is orthogonal to long-tail benchmarks (MINTQA,
CR-LT-KGQA), which retain question--path lexical alignment: rare
entities or near-synonymous relation paraphrases (e.g.,
\emph{hometown}$\leftrightarrow$\emph{place of birth}) only narrow
the lexical gap, leaving it well within reach of standard embedding
retrievers. \bench instead disallows relation labels and
intermediate entities from appearing on the question's surface,
widening the gap beyond what paraphrase can bridge,
and extends reasoning depth to 6 hops, beyond the 3--4 hop ceiling
of prior multi-hop KGQA benchmarks.
We describe how \bench operationalizes this restriction in \S\ref{sec:bench:pipeline}.


\begin{figure}[t]
\centering

\begin{minipage}[b]{0.30\textwidth}
  \centering
  \includegraphics[width=\linewidth]{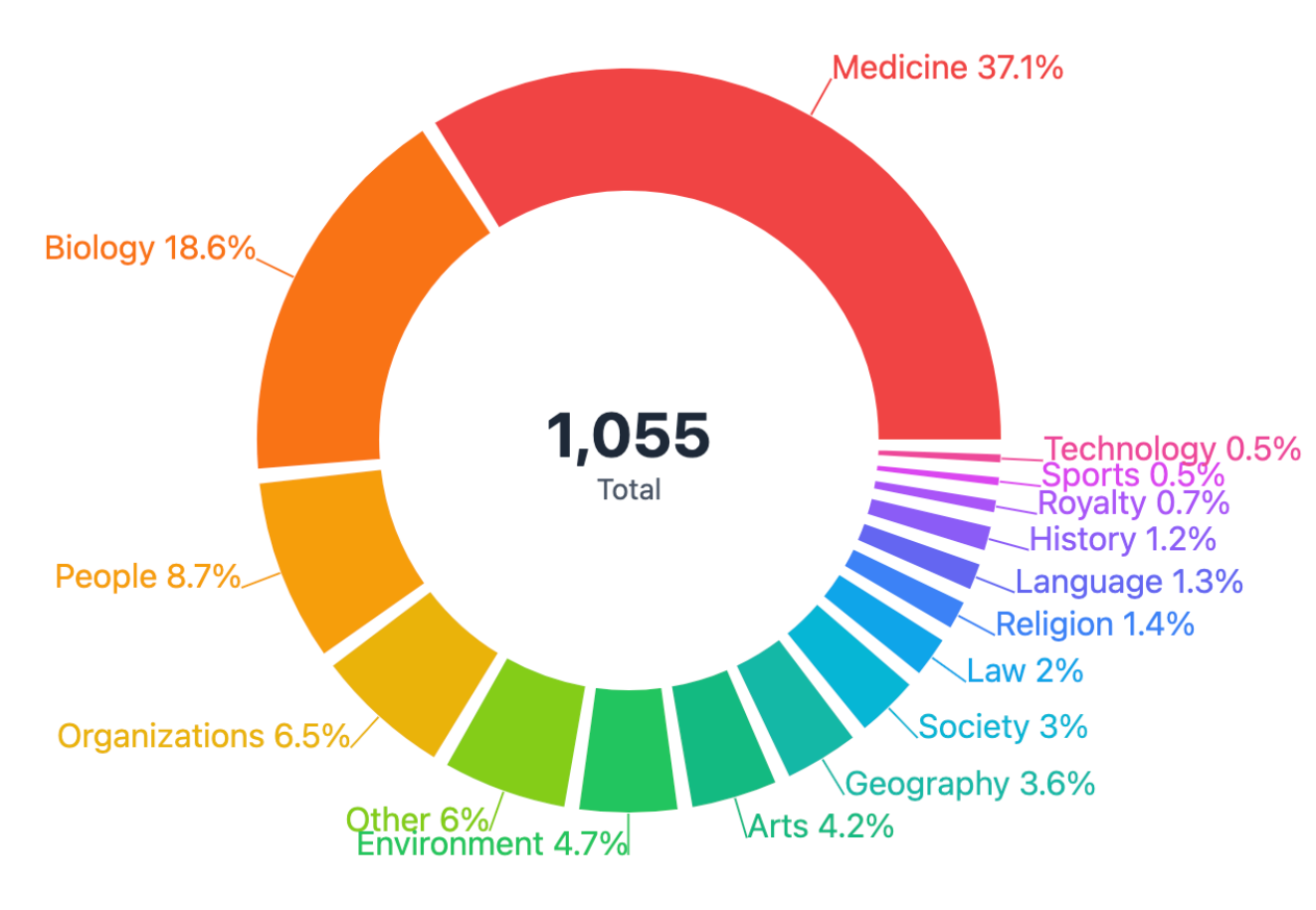}
  \subcaption{Domain distribution}
  \label{fig:domain_dist}
\end{minipage}\hfill
\begin{minipage}[b]{0.30\textwidth}
  \centering
  \includegraphics[width=\linewidth]{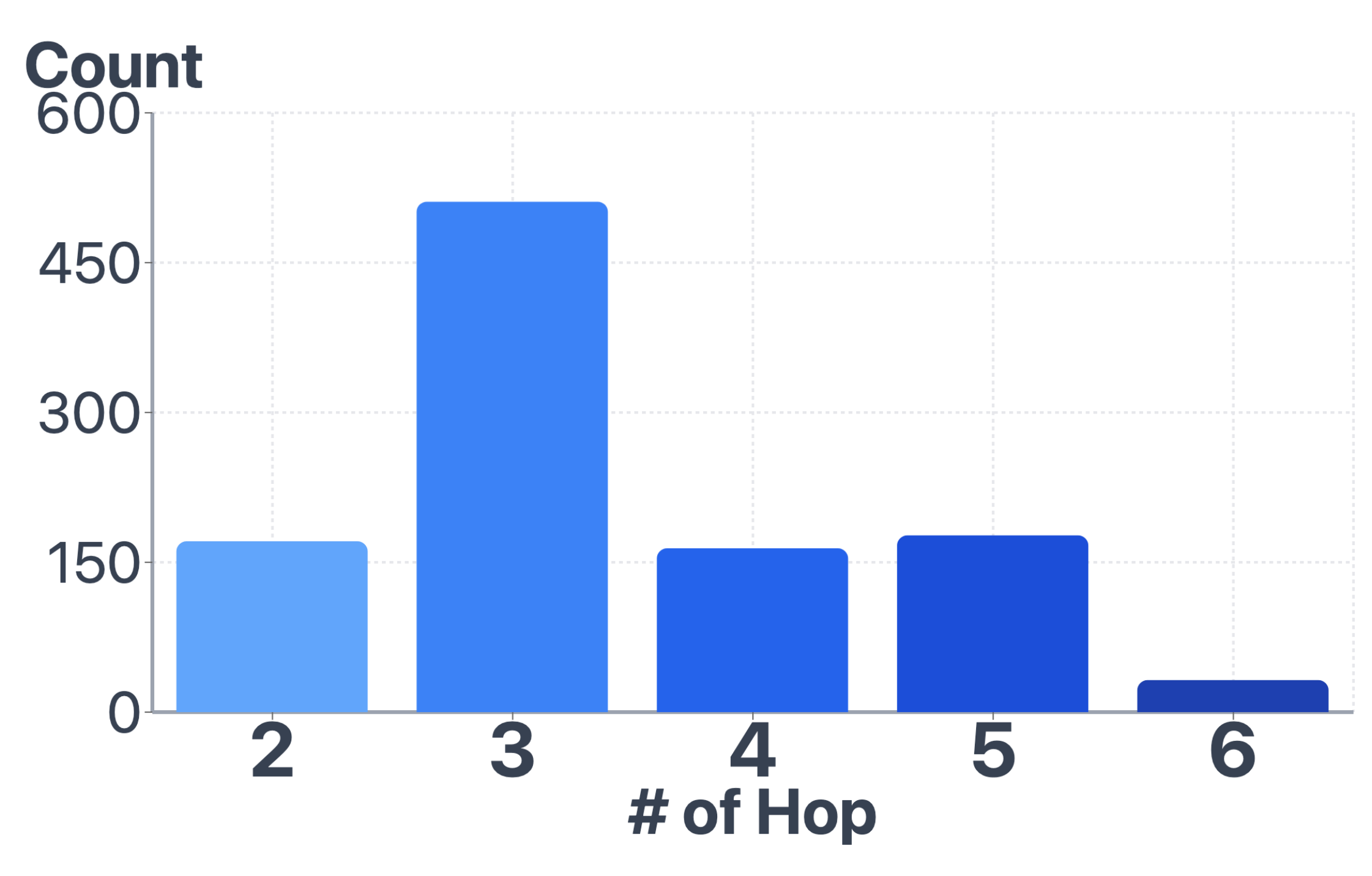}
  \subcaption{Hop distribution}
  \label{fig:hop_dist}
\end{minipage}\hfill
\begin{minipage}[b]{0.36\textwidth}
  \centering
  \footnotesize
  \setlength{\tabcolsep}{6pt}
  \begin{tabular}{lr}
    \toprule
    \textbf{Type} & \textbf{Count} \\
    \midrule
    Compressed-only   & 644 \\
    Cause-and-effect  & 250 \\
    Intermediate-node & 161 \\
    \midrule
    \textbf{Total}    & \textbf{1{,}055} \\
    \bottomrule
  \end{tabular}
  \subcaption{Question type distribution}
  \label{tab:type_dist}
\end{minipage}

\caption{Dataset statistics of \bench }
\label{fig:dataset_stats}
\end{figure}

\section{The \bench Benchmark}
\label{sec:bench}

\subsection{Preliminaries}
\label{sec:preliminaries}

We use Wikidata~\citep{wikidata} as our backbone KG. A KG is a
directed labeled graph $G=(V,E)$ in which $V$ is a set of
\emph{entities} and $E\subseteq V\times\mathcal{R}\times V$ is a
set of \emph{triples} $(h,r,t)$, where $h,t\in V$ are the head and
tail entities and $r\in\mathcal{R}$ is a \emph{relation} from a
fixed relation vocabulary $\mathcal{R}$. An \emph{$d$-hop path} $p$
is a sequence of $d$ consecutive triples
$e_0\xrightarrow{r_1}\cdots\xrightarrow{r_d}e_d$ with all
$(e_{i-1},r_i,e_i)\in E$; we call $(r_1,\ldots,r_d)$ its
\emph{relation chain} and $e_1,\ldots,e_{d-1}$ its \emph{intermediate
entities}. We use \emph{node} and \emph{entity} interchangeably and
consider only \emph{simple} paths (no repeated entity) throughout.
A \emph{topic entity} is an entity explicitly named in the
question~$Q$.

\begin{table*}[!htbp]
\centering
\footnotesize
\renewcommand{\arraystretch}{1.1}
\setlength{\tabcolsep}{6pt}
\setlength{\aboverulesep}{0.3ex}
\setlength{\belowrulesep}{0.3ex}
\begin{tabular}{@{}
   >{\centering\arraybackslash}m{0.13\textwidth}
   >{\raggedright\arraybackslash}m{0.59\textwidth}
   >{\raggedright\arraybackslash\scriptsize}m{0.21\textwidth} @{}}
\toprule
\textbf{Type} & \textbf{Question / Reasoning Path} & \textbf{Choices (a,b,c,d,e)} \\
\midrule
\multirow{2}{*}{\makecell{\textbf{(a)}\\ Compressed-only}}
 & \Qq{Betaxolol hydrochloride can treat which disease?} \newline
   {\scriptsize\Path{betaxolol HCl $\to$(physically interacts with)$\to$
   adrenoceptor $\beta$1 $\to$(encoded by)$\to$ ADRB1
   $\to$(genetic association)$\to$ arterial hypertension.}}
 & \dots \newline
   (d)~keratosis follicularis \newline
   \ans{(e)~None} \\
\cmidrule(l){2-3}
 & \Qq{If MBNL1 is altered, which gender could be affected more
   severely: men or women?} \newline
   {\scriptsize\Path{MBNL1 $\to$(genetic association)$\to$ myotonic dystrophy
   $\to$(subclass of)$\to$ male reproductive system disease.}}
 & (a)~women \newline
   \ans{(b)~men} \newline
   (c)~\dots \\
\midrule
\multirow{2}{*}{\makecell{\textbf{(b)}\\ Cause-and-effect}}
 & \textbf{(\textsc{Narrative})}~\Qq{Through which environmental
   issue do greenhouse gases most plausibly lead to a social
   issue?} \newline
   {\scriptsize\Path{greenhouse gas $\to$(has effect)$\to$ global warming
   $\to$(instance of)$\to$ environmental issue
   $\to$(subclass of)$\to$ social issue.}}
 & (a)~\dots human health\dots \newline
   \ans{(b)~Greenhouse gas emissions have effect\dots} \newline
   (c)~\dots disease outbreaks\dots \\
\cmidrule(l){2-3}
 & \textbf{(\textsc{Counterfactual})}~\Qq{If Robert of Beverley
   had never existed, which Catholic church building would most
   plausibly be missing or delayed?} \newline
   {\scriptsize\Path{R.\ of Beverley $\to$(notable work)$\to$ Westminster
   Abbey $\to$(instance of)$\to$ collegiate church
   $\to$(subclass of)$\to$ Catholic church bldg.}}
 & (a)~\dots \newline
   \ans{(b)~Westminster Abbey} \newline
   (c)~\dots \newline
   (d)~conventual church \newline
   (e)~\dots \\
\midrule
\makecell{\textbf{(c)}\\ Intermediate-node}
 & \Qq{Both Francesco Salvolini and Gustav Seyffarth are
   associated with which academic discipline?} \newline
   {\scriptsize\Path{Francesco Salvolini $\to$(student of)$\to$
   J.-F.~Champollion $\to$(field of work)$\to$ Egyptology;
   \quad Gustav Seyffarth $\to$(field of work)$\to$ Egyptology.}}
 & (a)~\dots \newline
   \ans{(b)~Egyptology} \newline
   (d)~sociolinguistics \newline
   (e)~None \\
\bottomrule
\end{tabular}
\caption{Representative examples of \bench three
question types. The correct option is highlighted in \ans{green};
\textcolor{pathblue}{gold paths} are shown in blue.}
\label{tab:question-types}
\end{table*}
\subsection{Construction pipeline}
\label{sec:bench:pipeline}

\bench keeps the standard path$\to$question construction direction
but inserts a \emph{compressibility filter} between the two: for each
candidate path, we first ask whether the multi-hop chain admits a
compressed lay phrasing in which the path's relations no longer
surface as tokens. The filter operates in two stages: an LLM-based
feasibility check screens candidate paths
(GPT-5-mini~\citep{gpt5}); the LLM then drafts a compressed
question for each surviving path, and the authors manually verify
the resulting (path, question) pair before it enters the released
benchmark. As a result, the released questions are not paths
converted under a generation prompt, but paths \emph{selected for}
their amenability to compression and items \emph{verified for}
faithful compressed phrasing.

\textbf{Step 1 — Domain selection.}
We select various domains including anatomy, disease, law, and
people (Figure~\ref{fig:dataset_stats}).  Each domain satisfies
two criteria: (i)~it admits a sufficient number of type-stable
Wikidata entities; (ii)~the domain has a well-defined set of
characteristic predicates that we can enumerate up front (e.g.,
\emph{mother}, \emph{spouse}, \emph{educated at} for the people
domain). 

\textbf{Step 2 — Seed entity \& path enumeration.}
We sample approximately 700 seed entities per domain and enumerate
simple paths of hop 2--6 from each seed. This yields 85{,}320
candidate paths.

\textbf{Step 3 — Compressibility check \& path-to-question
generation.}
For each seed entity, we route two pools of paths through the
pipeline: (i)~the top-20 paths whose relation sequences are rarest
across the full path corpus, to encourage diverse question phrasings,
and (ii)~100 additionally sampled paths drawn uniformly at random,
for broad coverage.  Each candidate path $p$ is processed in two
sub-steps:
\begin{description}[nosep, leftmargin=*]
  \item[(3a) LLM-based compressibility check.]  We prompt\footnote{Details in appendix~\ref{app:questionGen} to check compressibility.} the LLM to decide whether $p$ admits a compressed lay phrasing in which
    the path's relations are not lexically present in the question.
    Paths that the model judges non-compressible are discarded at
    this stage.
  \item[(3b) Question generation.]  For each path that passes the
    compressibility check, the LLM generates one compressed
    natural-language question. This step produces approximately 7{,}600 candidate questions.
\end{description}

\textbf{Step 4 — Manual verification.}
The authors---AI PhD researchers familiar with the relevant domain
ontologies---review every candidate against three criteria(C1-3), which
together define what counts as a valid \bench instance. After this
step, 1{,}055 of 7{,}615 candidates are retained.
\begin{description}[nosep, leftmargin=*]
  \item [(C1) Restricted path--question alignment.]
    The path's relations must not be recoverable from the question
    by lexical or near-synonymous matching, so that answering
    requires genuine multi-hop reasoning. We re-check the LLM
    judgment from Step~3a manually. For example, in the question
    \emph{``Which disease could Ludwig von der Pfalz be genetically
    predisposed to through family history?''}, the gold path
    contains \emph{father}, \emph{mother}, \emph{medical condition},
    and \emph{subclass of}---none of which appears as a token; the
    entire 5-hop chain is compressed into the single phrase
    ``family history.''
  \item [(C2) Multi-hop guarantee (no 1-hop shortcut).]
    Every gold path is 2--6 hops, and we discard any question for
    which a 1-hop relation between the same entity pair already
    exists in Wikidata. For example, for \emph{``What disease can
    thalidomide help with?''} the intended multi-hop path is
    \emph{thalidomide $\to$(physically interacts with)$\to$Cereblon
    $\to$(encoded by)$\to$CRBN
    $\to$(genetic association)$\to$multiple myeloma}, but a 1-hop
    \emph{thalidomide $\to$(medical condition treated)$\to$multiple
    myeloma} edge exists, so the question is rejected.
  \item [(C3) Question--path semantic alignment.]
    The compressed phrasing must accurately summarize the path's
    meaning in fluent, natural-sounding language. When a path passes
    (C1) and (C2) but the LLM-drafted question reads awkwardly or
    under-specifies the intent, the authors lightly edit the question
    to improve fluency while preserving its compressed phrasing
    (i.e., without reintroducing tokens from the path's relations).
\end{description}

\textbf{Step 5 — Distractor generation.}
For each retained question we generate three content distractors
plus a fixed (e)~\texttt{None} option. Distractor entities are
chosen so that surface-level shortcuts are insufficient: they share
the entity type of the gold answer and lie on the gold or adjacent
paths. For questions whose options are sentence-form causal
narratives, we relax type matching and prompt an LLM with the gold
path and selected distractor entities to synthesize full causal
chains rather than single-entity swaps. A subset of questions is
additionally converted into \texttt{None}-answer items by replacing
the gold entity in the candidate set with an additional type-matched
distractor, leaving the question and gold path intact\footnote{Distractor Generatin details in Appendix~\ref{app:distractors}}. 

\subsection{Question types}
\label{sec:bench:types}

\bench defines three question types, each capturing a distinct mode of
path--question alignment collapse:
\textbf{(a) compressed-only}, where an N-hop relation chain is lexically summarized into one or two surface tokens;
\textbf{(b) cause-and-effect}, where the question requires the \emph{interpretation} of the multi-hop chain rather than retrieval of any single node — either a narration of the path or a counterfactual reading of what would not exist absent some entity; and
\textbf{(c) intermediate-node}, where the question gives two topic entities and asks how they are connected, requiring the model to uncover an unstated bridging concept that the question itself does not name. Table~\ref{tab:question-types} shows the examples\footnote{The benchmark is derived from Wikidata, which is available under CC0 1.0. We release \bench under CC BY 4.0.}.

\subsection{Holistic Path Selection (Our Method)}
\label{sec:method:holistic}
Rather than prune hop-by-hop KG exploration or rely on embedding-based
retrieval, our method presents \emph{all candidate relation chains}
reachable from the topic entity to the LLM at once and lets the
model select the answer chain holistically.
Let $G_q$ denote the per-question local subgraph induced by the
topic entity (\S\ref{sec:bench:format}),
$\mathbf{r}=(r_1,\ldots,r_d)$ the relation chain of a $d$-hop path
in $G_q$, $\mathcal{L}_t\subseteq\mathcal{R}$ the relations
available at $e_{t-1}$, and $\mathcal{C}$ the set of all candidate
relation chains in $G_q$ reachable from the topic entity. hop-by-hop KG exploration factorizes the joint, while our method estimates it
directly:
\[
\underbrace{P(\mathbf{r}\mid Q, G_q)\approx\prod_{t=1}^{d}P(r_t\mid Q, e_{t-1}, \mathcal{L}_t)}_{\text{hop-by-hop}}
\quad\text{vs.}\quad
\underbrace{P(\mathbf{r}\mid Q, G_q)\approx P(\mathbf{r}\mid Q, \mathcal{C})}_{\text{holistic (ours)}}.
\]
This trades hop-level cues for whole-against-whole alignment between
the question and the full chain, eliminating the early commitments
that propagate errors in stepwise pruning. It also bypasses the
embedding-based retrieval bottleneck by presenting chains as plain
text to an off-the-shelf LLM, requiring no KG-specific training.


\subsection{A Three-Stage Pipeline of Holistic Path Selection}
\label{sec:method:pipeline}

We realize the holistic objective through three LLM calls operating
at increasing levels of grounding. Let
$\mathcal{P}_{\mathbf{r}}\!=\!\{p\!\in\! G_q : \mathrm{rel}(p)_{1:|\mathbf{r}|}\!=\!\mathbf{r}\}$
denote the paths in $G_q$ whose relation prefix matches $\mathbf{r}$,
and $\tau:V\!\to\!\mathcal{T}$ a type lookup, where $\mathcal{T}$ is
the set of entity types. Hyperparameters $K_1$ and $K_2$ bound the
number of chains and paths retained at Steps~1 and~2,
respectively.

\paragraph{Step~1 -- Schema-level chain selection.}
Each candidate chain $\mathbf{r}\!\in\!\mathcal{C}$ at depth $d$ is
shown to the LLM under the \emph{topic-typed} serialization
\begin{equation}
\sigma_1(\mathbf{r}) \;=\; e_0{:}[\tau(e_0)] \xrightarrow{r_1} [\hat{\tau}_1] \xrightarrow{r_2} \cdots \xrightarrow{r_d} [\hat{\tau}_d],
\label{eq:sigma1}
\end{equation}
where $\hat{\tau}_i$ is the modal entity type at hop $i$ across
$\mathcal{P}_{\mathbf{r}}$. Only the topic $e_0$ is grounded by
surface form; downstream slots reveal types only, so the model
commits to chains by relational structure rather than guessing
surface forms. The LLM returns
\[
\mathcal{C}^{*} \;=\; \mathrm{LLM}_1\bigl(Q,\,\{\sigma_1(\mathbf{r})\}_{\mathbf{r}\in\mathcal{C}}\bigr)\subseteq\mathcal{C},
\qquad |\mathcal{C}^{*}|\le K_1.
\]

\paragraph{Step~2 -- Grounded path selection.}
Selected chains are expanded to grounded paths
$\mathcal{P}^{*}\!=\!\bigcup_{\mathbf{r}\in\mathcal{C}^{*}}\mathcal{P}_{\mathbf{r}}$
and re-serialized with full instantiation
\begin{equation}
\sigma_2(p) \;=\; e_0{:}[\tau(e_0)] \xrightarrow{r_1} e_1{:}[\tau(e_1)] \cdots \xrightarrow{r_d} e_d{:}[\tau(e_d)].
\label{eq:sigma2}
\end{equation}
The LLM selects
\[
\mathcal{P}^{**} \;=\; \mathrm{LLM}_2\bigl(Q,\,\{\sigma_2(p)\}_{p\in\mathcal{P}^{*}}\bigr),
\qquad |\mathcal{P}^{**}|\le K_2,
\]

\paragraph{Step~3 -- Answer commitment.}
At this final stage the LLM additionally observes the answer choices
$\mathcal{A}=\{a,b,c,d,e\}$, together with the question $Q$ and the
grounded paths from Step~2, and commits to one option:
\begin{equation}
\hat{a} \;=\; \mathrm{LLM}_3\bigl(Q,\,\mathcal{A},\,\{\sigma_2(p):p\in\mathcal{P}^{**}\}\bigr).
\label{eq:step3}
\end{equation}
The two serializations $\sigma_1,\sigma_2$ induce a deliberate
\emph{information hierarchy}: Steps~1 and~2 see only the question
and the KG paths, and $\mathcal{A}$ enters the pipeline only at
Step~3. Each step's role---chain selection, instance filtering, and
answer decision---is thus structurally separated, and retrieval
cannot be steered by surface cues from the answer options. 

\paragraph{Two scheduling variants.}
We pair this three-stage backbone with two outer loops differing
primarily in how $\mathcal{C}$ is exposed to Step~1. \onec{} feeds
the entire $\mathcal{C}$ to Step~1 in one shot, regardless of
depth, and retries the full pipeline up to $T$ times whenever
Step~3 emits \texttt{unsure}; each retry excludes chains selected
in previous attempts. \sbd{} instead runs the pipeline in a depth
loop $d=2,\dots,D$, exposing only the depth-$d$ chain set
$\mathcal{C}_d=\{\mathbf{r}\in\mathcal{C}:|\mathbf{r}|=d\}$ at each
depth, advancing to $d{+}1$ on \texttt{unsure} and treating any of
\texttt{a}--\texttt{e} as a commit at the current depth\footnote{Full
pseudocode is in Appendix~\ref{app:algorithms}.}.

\section{Experiments}
\subsection{Evaluation protocol}
\label{sec:bench:eval}

\begin{equation*}
\text{Acc.} = \tfrac{\#\text{correct}}{\#\text{total}},\quad
\text{AR} = \tfrac{\#\text{answered}}{\#\text{total}},\quad
\text{SA} = \tfrac{\#\text{correct}}{\#\text{answered}},\quad
\#\text{total} = \#\text{answered} + \#\text{unanswered}.
\end{equation*}
\paragraph{Metrics.}
We report three metrics:
Accuracy(AC) is reported for every method. AnswerRate(AR) and SelectiveAccuracy(SA) are only meaningful
for methods that support an explicit \emph{abstention
mechanism}---i.e., methods that can deliberately refuse to answer(unanswered) when
their internal evidence is insufficient, which in our evaluation applies to R2-KG~\citep{r2kg} and Holistic Path Selection.

\paragraph{Per-question subgraphs.}
\label{sec:bench:format}
\bench provides topic entities for every question, so all methods
begin exploration from the topic-entity neighborhood; nodes
unreachable within a reasonable hop budget cannot participate in
the gold path or in any retrieval trajectory we observed. We
therefore evaluate all methods on a pre-extracted per-question
subgraph $G_q \subseteq G$, anchored at the topic entities of $q$
and guaranteed to contain the gold reasoning path; the union of
these subgraphs covers roughly 68K entities. This protocol is
consistent with recent KGQA benchmarks~\citep{crltkgqa,
dynamickgqa, kgqagen} and methods~\citep{rog, fidelis,
pog}\footnote{Subgraph extraction details are in
Appendix~\ref{app:subgraph}.}.

\paragraph{Path-conditioned upper bound.}
We additionally report the accuracy obtained when the gold path is provided
to the LLM \emph{together with} the question. This quantifies the dataset's
ambiguity ceiling: any deviation from 100\% reflects either dataset noise
or the LLM's residual reasoning limits even when given the answer path.
 
\paragraph{Pseudoword analysis (dataset self-containedness check).}
\label{sec:bench:pseudoword}
To verify that \bench is solvable from KG structure alone, we
additionally evaluate all methods on a \emph{pseudoword variant} of
the benchmark, in which every non-type entity surface form(both for $G_q$ and $Q$) is
replaced with a meaningless pseudo-token,
while \emph{type entities} (those appearing as the object of
\texttt{instance of} or \texttt{subclass of} edges) and all
relation labels are kept intact. For instance, \textit{Original-split}:
\textsc{thalidomide}\,$\xrightarrow{\text{instance of}}$\,
\textsc{stereoisomers mixture}
becomes
\textit{Pseudoword-split}:\texttt{E\_0321}\,$\xrightarrow{\text{instance of}}$\,
\textsc{stereoisomers mixture}.
This preserves the schema-level information needed to identify
\emph{what kind of thing} each entity is, while removing
identifying surface cues that an LLM could match against external
parametric memory.

\paragraph{Experimental Setting.}
\label{sec:resource}
All methods, including baselines and ours, call the same two LLMs through
an identical chat-completion interface: (1)~\textbf{GPT-5-mini}~\citep{gpt5}
with \texttt{reasoning\_effort=minimal} and \texttt{verbosity=low}, and
(2)~\textbf{Qwen3-32B}~\citep{qwen3} with \texttt{enable\_thinking=False}
to ensure fair comparison across baselines. For text embeddings,
KAPING~\citep{kaping} and KARPA~\citep{karpa} use OpenAI's
\texttt{text-embedding-3-small}. All open-model experiments are
conducted on $4\times$NVIDIA A100 80GB GPUs\footnote{Check Appendix~\ref{app:resource} for resource detail}. We set $T=2, D=6$ for \onec and \sbd.
For every baseline we adopt the official implementation released by the
original authors and keep both the algorithm and the prompts essentially
unchanged. We make two minimal modifications to fit our benchmark.
First, since \bench contains gold paths of up to 6 hops, we extend the
default beam-search depth of baselines (ToG, PoG,
FiDeLiS, KARPA, KAPING, R2-KG) to 6 so that no gold path is structurally
unreachable; all other hyperparameters (beam width, top-$k$, etc.)
follow the original papers. Second, several baselines ship with
few-shot exemplars written in Freebase\citep{freebase} notation; we rewrite
these exemplars to the Wikidata\citep{wikidata} surface form used in our backbone KG,
preserving the original prompt structure and reasoning style.
All KG methods (baselines and \onec{}/\sbd{}) operate choice-blind:
$\mathcal{A}$ is revealed only at the final answer-commitment step,
preventing answer-conditioned shortcuts from leaking into retrieval.
\textsc{ZeroShot} and \textsc{CoT}~\citep{cot} receive $\mathcal{A}$
upfront, since they have no retrieval stage.

\begin{table}[t]
\centering
\footnotesize
\setlength{\tabcolsep}{5pt}
\renewcommand{\arraystretch}{0.95}
\setlength{\aboverulesep}{0.3ex}
\setlength{\belowrulesep}{0.3ex}
\begin{tabular}{ll cc cc}
\toprule
& & \multicolumn{2}{c}{\textbf{Qwen3-32B}} & \multicolumn{2}{c}{\textbf{GPT-5-mini}} \\
\cmidrule(lr){3-4}\cmidrule(lr){5-6}
\textbf{Paradigm} & \textbf{Method}
  & Original & Pseudoword & Original & Pseudoword \\
\midrule
\multirow{2}{*}{Closed-book}
  & Zero-shot     & 63.8 & 25.2 & 66.6 & 25.7 \\
  & CoT~\citep{cot}           & 64.6 & 25.3 & 69.1 & 23.1 \\
\midrule
Plan-then-retrieve
  & KARPA~\citep{karpa}         & 40.1 & 28.0 & 44.3 & 36.0 \\
\midrule
Similarity retrieval
  & KAPING~\cite{kaping}        & 69.1$^{\dagger}$ & 45.4 & 70.1$^{\dagger}$ & 41.7$^{\dagger}$ \\
\midrule
\multirow{4}{*}{hop-by-hop KG exploration}
  & ToG~\cite{tog}           & 50.2 & 29.0 & 55.1 & 32.4 \\
  & PoG~\cite{pog}           & 66.4 & 47.3 & 65.7 & 37.7 \\
  & FiDeLiS~\cite{fidelis}       & 42.7 & 31.1 & 46.6 & 34.9 \\
  & R2-KG~\cite{r2kg}         & 64.7 & 47.5$^{\dagger}$ & 66.6 & 40.9 \\
\midrule
\rowcolor{cyan!15}\cellcolor{cyan!15}
  & \sbd   & \underline{74.9} & \underline{56.0} & \textbf{82.5} & \textbf{59.6} \\
\rowcolor{cyan!15}\cellcolor{cyan!15}\multirow{-2}{*}{Holistic Path Selection(\textbf{Ours})}
  & \onec  & \textbf{78.5} & \textbf{56.3} & \underline{74.0} & \underline{51.5} \\
\midrule
\rowcolor{yellow!25}\cellcolor{yellow!25}Oracle & UpperBound & 96.5 & 91.7 & 97.9 & 90.1 \\
\bottomrule
\end{tabular}
\caption{Accuracy on \bench for both \textit{Original} split and \textit{Pseudoword} split.
\textbf{Bold} = best, \underline{underline} = second-best,
$^{\dagger}$ = third-best per column (excluding the Oracle).}
\label{tab:main}
\end{table}

\subsection{Results}
\label{sec:results:main}


\paragraph{The Pseudoword split removes entity-name shortcuts.}
Table~\ref{tab:main} reports accuracy on \bench{} across both backbones
and both splits(Original vs. Pseudoword). Closed-book LLMs (Zero-shot, CoT) achieve $63.8$--$69.1\%$ accuracy on
the Original split but collapse to $23.1$--$25.7\%$ on Pseudoword,
indistinguishable from a trivial baseline that ignores the question
entirely and always picks the most frequent answer letter ('b',
$26.0\%$ of the data). This $\sim$$40$-point drop, observed
consistently across both backbones, confirms that closed-book
performance on Original is largely driven by memorized Wikidata content
rather than reasoning. We therefore treat Pseudoword as the primary
diagnostic split: it isolates KG-grounded reasoning from parametric
priors and, for the same reason, is the split on which the
paradigm-level comparisons below are most informative.

\paragraph{All three paradigms degrade under alignment collapse,
with distinct severity.}
\label{res:failureofbaseline}
§\ref{sec:intro:hidden} pointed out that the three paradigms share a
single underlying failure cause but differ in \emph{where} the
failure manifests. Table~\ref{tab:main} shows that the Original-split numbers support this: three
baselines spanning two paradigms fall \emph{below} closed-book CoT
($64.6/69.1$)---FiDeLiS ($42.7/46.6$) and ToG ($50.2/55.1$) from
hop-by-hop, and KARPA ($40.1/44.3$) from plan-then-retrieve, on
Qwen/GPT---while the remaining baselines (PoG, R2-KG, KAPING) only
match or slightly exceed it. A plausible reading is that retrieval
errors in the underperforming methods override correct parametric
memories, while verifier-augmented or globally-scoped retrieval is
more robust to such interference.
The Pseudoword split disentangles this from memorization. Once
entity-name shortcuts are removed, all six KG-augmented baselines
exceed closed-book CoT, yet the per-paradigm best on Pseudoword
(GPT) remains tightly clustered---R2-KG $40.9$ (hop-by-hop), KARPA
$36.0$ (plan-then-retrieve), KAPING $41.7$ (similarity retrieval),
within a 5.7-point band and all $\geq 17.9$ points below \sbd{}. Two implications follow. First,
KG benefit is real for every paradigm under alignment collapse, but
its magnitude depends more on retrieval/verification strategy than
on paradigm category. Second, no prior paradigm escapes the
alignment-collapse ceiling\footnote{Case study in
Appendix~\ref{app:qualtative_anal}}.

\begin{table}[t]
\centering
\footnotesize
\setlength{\tabcolsep}{3pt}                
\renewcommand{\arraystretch}{0.95}         
\setlength{\aboverulesep}{0.3ex}
\setlength{\belowrulesep}{0.3ex}
\begin{tabular}{l ccc ccc ccc ccc}
\toprule
& \multicolumn{6}{c}{\textbf{Qwen3-32B}} & \multicolumn{6}{c}{\textbf{GPT-5-mini}} \\
\cmidrule(lr){2-7}\cmidrule(lr){8-13}
& \multicolumn{3}{c}{Original} & \multicolumn{3}{c}{Pseudoword}
& \multicolumn{3}{c}{Original} & \multicolumn{3}{c}{Pseudoword} \\
\cmidrule(lr){2-4}\cmidrule(lr){5-7}\cmidrule(lr){8-10}\cmidrule(lr){11-13}
\textbf{Method}
  & Acc & AR & SA & Acc & AR & SA
  & Acc & AR & SA & Acc & AR & SA \\
\midrule
R2-KG          & 64.7$^{\dagger}$ & 85.6$^{\dagger}$ & 75.6$^{\dagger}$
               & 47.5$^{\dagger}$ & \underline{94.9} & 50.0$^{\dagger}$
               & 66.6$^{\dagger}$ & \underline{96.8} & 68.9$^{\dagger}$
               & 40.9$^{\dagger}$ & \underline{97.7} & 41.8$^{\dagger}$ \\
\midrule
\rowcolor{cyan!15}
\sbd    & \underline{74.9} & \underline{97.2} & \underline{77.1}
        & \underline{56.0} & 91.9$^{\dagger}$ & \textbf{60.9}
        & \textbf{82.5} & 96.6$^{\dagger}$ & \textbf{85.4}
        & \textbf{59.6} & 92.2$^{\dagger}$ & \textbf{64.6} \\
\rowcolor{cyan!15}
\onec   & \textbf{78.5} & \textbf{99.4} & \textbf{78.9}
        & \textbf{56.3} & \textbf{99.5} & \underline{56.6}
        & \underline{74.0} & \textbf{99.8} & \underline{74.2}
        & \underline{51.5} & \textbf{100.0} & \underline{51.5} \\
\bottomrule
\end{tabular}
\caption{Accuracy ($Acc$), answer rate ($AR$), selective accuracy
($SA$) for methods with explicit abstention mechanisms on \bench and R2-KG.
\textbf{Bold} = best, \underline{underline} = second-best,
$^{\dagger}$ = third-best per column.}
\label{tab:abstain}
\end{table}

\paragraph{Holistic Path Selection dominates and exposes a complementary trade-off.}
\label{res:hopbypruning_prob}
\sbd{} and \onec{} take both the best and the second-best slot in
every column of Table~\ref{tab:main}, with substantial margins over
the strongest competing baseline; the gap is largest on Pseudoword,
where entity-name shortcuts are removed. Under alignment collapse,
no local fragment of the topic-entity neighborhood (a single
relation, a question-only plan, or a top-$k$ slice of triples)
carries enough signal to identify the gold path, whereas presenting
the $d$-hop neighborhood jointly to the LLM lets it commit globally
to the chain that fits the question. KAPING is the only baseline
that also views the subgraph globally, yet the gap between
path-level selection (ours) and triple-level similarity ranking
(KAPING) widens further on Pseudoword.

\paragraph{Further Analysis; Holistic variants recover the gold path more reliably.}
A natural question is whether Holistic Path Selection succeeds by genuinely
recovering the gold KG path, or merely by arriving at the correct answer
through other means. To investigate this, Table~\ref{tab:path_retrieval}
reports two path-level metrics: \textbf{Path~F1}, the F1 score between the
retrieved and gold relation paths, and \textbf{Answer-Correct Path~F1}
(AC-Path~F1), the same score restricted to correctly answered examples.
Both Holistic variants outperform all baselines on both metrics across both
backbones, with Path~F1 gains of $17$--$21$ points over the strongest
baseline (R2-KG). AC-Path~F1 exceeds overall Path~F1 by $7$--$9$ points
for Holistic methods but only $3$--$6$ points for baselines, indicating
that correct answers by Holistic methods are more strongly coupled to
successful gold-path recovery. Although Path~F1 and answer accuracy do not
induce identical rankings across methods, they exhibit the same overall
tendency: more accurate path retrieval is associated with more reliable
answer commitment.
\begin{table}[t]
\centering
\footnotesize
\setlength{\tabcolsep}{5pt}
\renewcommand{\arraystretch}{0.95}
\setlength{\aboverulesep}{0.3ex}
\setlength{\belowrulesep}{0.3ex}
\begin{tabular}{l cc cc}
\toprule
& \multicolumn{2}{c}{\textbf{Qwen3-32B}} & \multicolumn{2}{c}{\textbf{GPT-5-mini}} \\
\cmidrule(lr){2-3}\cmidrule(lr){4-5}
\textbf{Method}
  & Path F1 & AC-Path F1 & Path F1 & AC-Path F1 \\
\midrule
KARPA   & 28.6 & 31.9 & 35.4 & 39.6 \\
KAPING  & 43.8 & 47.2 & 41.7 & 45.5 \\
ToG     & 31.7 & 35.8 & 34.9 & 39.3 \\
PoG     & 45.9 & 50.0 & 40.5 & 44.7 \\
FiDeLiS & 32.9 & 36.6 & 34.0 & 38.1 \\
R2-KG   & 49.2 & 54.8 & 45.8 & 51.7 \\
\midrule
\rowcolor{cyan!15} \sbd{}      & \underline{62.8} & \underline{71.5} & \textbf{69.6} & \textbf{78.2} \\
\rowcolor{cyan!15} \onec{}     & \textbf{66.9} & \textbf{74.8} & \underline{60.8} & \underline{68.1} \\
\bottomrule
\end{tabular}
\caption{Path-retrieval performance on \bench{}.
AC-Path F1 = Answer-Correct Path F1, computed only over correctly answered examples.
\textbf{Bold} = best, \underline{underline} = second-best.}
\label{tab:path_retrieval}
\end{table}

\paragraph{The two variants trade off AR against SA}
In the Table~\ref{tab:abstain}, the two variants realize this joint view differently and thus
trade Answer Rate(AR) against Selective Accuracy(SA). \sbd{} grows the visible chain set one depth
at a time \emph{without pruning earlier hops}: at each depth the
LLM sees all chains up to $d$ hops and decides whether they
suffice (\texttt{commit}) or one more depth is needed (\texttt{unsure}).
The depth schedule is a presentation choice, not a pruning step.
\onec{} instead exposes the full $N$-hop neighborhood at once,
yielding higher AR but admitting chains beyond the gold hop count
as distractors. AR is consistently higher for \onec{}, while SA
favors \sbd{} in three of four settings.
The two are complementary: \onec{} suits deployments preferring a
committed answer for most queries, \sbd{} those where low-evidence
cases are better left unanswered.

\paragraph{A substantial headroom remains relative to the oracle.}
The Oracle row of Table~\ref{tab:main} reports accuracy when the
gold path is provided together with the question (96.5--97.9 on
Original, 90.1--91.7 on Pseudoword). Our best results
trail the oracle by 15.4--18.0 points on Original and
30.5--35.4 on Pseudoword. Since the oracle stays above 90\%
even on Pseudoword, the questions \emph{are} answerable from the KG
itself; what closes this gap, then, is not the data but better
retrieval and path-selection. We see one promising direction:
instead of ranking candidate paths by static
question--path similarity, score them by \emph{how useful each path
is for committing to an answer}.

\paragraph{Holistic Path Selection generalizes across benchmarks and KGs.}
To test whether the advantage of Holistic Path Selection is specific to
\bench{} or Wikidata, we evaluate on transformed samples from MetaQA
(WikiMovies) and ComplexWebQuestions (Freebase), where explicit
relation-level lexical cues are concealed while gold paths are preserved.
Across three LLM backbones (GPT-5-mini, Claude-Sonnet-5, Gemini-2.5-Pro),
FlatHolistic achieves $94$--$98\%$ on transformed MetaQA and $93$--$98\%$ on
transformed CWQ, while the strongest baselines decline by up to $24$ points
after transformation. Both Holistic variants degrade by only $0.7$--$3.7$
points on average, confirming that the robustness to alignment collapse
generalizes beyond \bench{} and Wikidata
(Appendix~\ref{app:generalization}).


\section{Conclusions and Limitations}
\label{sec:limitation}
We introduced \bench, a manually verified KGQA benchmark whose
surface forms do not leak the gold KG path, and \emph{Holistic
Path Selection}, a reasoning paradigm that selects the
answer chain globally rather than pruning hop by hop. Our \sbd
and \onec implementations consistently outperform strong KGQA
baselines, with the largest gains on Pseudoword; a side finding
is that all three prior paradigms fail in structurally related
ways under compressed questions---pointing to path--question
alignment as a missing evaluation axis.
\textbf{Limitations.} \bench is built on Wikidata only and
evaluated with two backbones, and its questions are
multiple-choice and English-only. The remaining gap to the
oracle suggests that scoring paths by \emph{how useful each is
for committing to an answer} is a promising direction.

\begin{ack}
This work was supported by the Institute of Information \& Communications Technology Planning \& Evaluation (IITP) grants (No.RS-2019-II190075, No.RS-2022-II220984), and the National Research Foundation of Korea (NRF) grant (NRF-2020H1D3A2A03100945, RS-2023-00262527), funded by the Korean government (MSIT).
\end{ack}

\bibliographystyle{abbrvnat}
\bibliography{references}


\appendix

\section{Per-Question Subgraph Extraction}
\label{app:subgraph}

For every question $q$ in \bench{}, we precompute a per-question
subgraph $G_q$ from Wikidata that all methods — baselines and ours
alike — share as the same KG substrate.  Centralising this extraction
yields two benefits.  First, querying the live Wikidata SPARQL
endpoint on every method call would be prohibitively slow and
introduce non-determinism (rate limits, transient failures) into
benchmark numbers.  Second, by giving every method the same $G_q$,
any performance gap is attributable to the method itself rather than
to differences in KG access or coverage, which is consistent with
recent KGQA practice~\citep{rog, fidelis, pog, crltkgqa, dynamickgqa,
kgqagen}.

\paragraph{Tree expansion from each topic entity.}
$G_q$ is built by expanding outward from each topic entity
$e_0\in\texttt{given\_entity\_qids}(q)$ via breadth-first SPARQL
queries up to a maximum depth of $H{=}6$ hops.  Each topic entity
yields its own tree; if a question has multiple topic entities, the
trees are constructed independently and their candidate paths are
unioned to form $G_q$.  Outgoing edges only are stored (the saved
graph is tree-structured); methods that need incoming relations
(e.g., ToG, PoG) build a reverse index on the fly.

At every node in the frontier, edge selection proceeds in two stages:
(i)~any edge that matches the gold path's relation and destination
at the current hop is \emph{always} included; (ii)~the remaining
budget is filled with non-gold edges sampled uniformly from the
node's outgoing edges, restricted by the per-hop branching schedule
detailed below.  Stage~(i) guarantees that the gold reasoning path
is present in $G_q$.  Stage~(ii) supplies plausible distractor
structure around it.  In the rare cases where natural expansion
fails to include the gold path (e.g., the SPARQL endpoint returns
incomplete results for an intermediate node), a post-processing
patch step forcibly inserts the gold path so that
\textbf{every $G_q$ is guaranteed to contain the gold path}.

\paragraph{Adaptive branching schedule.}
The number of edges retained per node decreases with depth:

\begin{center}
\small
\begin{tabular}{lcc}
\toprule
\textbf{Hop} & \textbf{max relations} & \textbf{max entities} \\
\midrule
0--1 & 5 & 2 \\
2--3 & 3 & 2 \\
4--5 & 2 & 1 \\
\bottomrule
\end{tabular}
\end{center}

This wide-to-narrow schedule is a deliberate difficulty design.
\emph{Path retrieval is fundamentally a sequential decision problem:
choices made near the topic entity propagate to and constrain every
subsequent hop.}  If the topic-entity neighborhood already contains
only a handful of relations, the retrieval task collapses into
trivial enumeration, and no method — ours or any baseline — can
demonstrate genuine reasoning ability because the search space is
nearly empty from the start.  We therefore allocate the largest
branching factor at the root (up to 10 outgoing edges) so that the
first hop forces methods to discriminate among many semantically
plausible directions, and we tighten the budget at deeper hops to
prevent combinatorial blow-up while leaving enough non-gold paths to
remain non-trivial.  In effect, the schedule shifts the difficulty
toward early-hop decisions, where the impact of a wrong choice is
greatest.  This makes \bench{} sensitive to the quality of
\emph{schema-level chain selection} — the very capability our method
targets at Step~1.

\paragraph{Type-triple collection.}
After tree expansion, every entity that appears anywhere in $G_q$
receives a single canonical type to support type-aware reasoning.
For each entity we issue one additional SPARQL query and choose: the
first \texttt{P31} (\textit{instance of}) value sorted by destination
QID for determinism, or — if \texttt{P31} is absent — the first
\texttt{P279} (\textit{subclass of}) value under the same ordering, or
a placeholder ``\texttt{?}'' if neither is available.  We retain only
one type per entity (rather than the full \texttt{P31} list) because
Wikidata frequently lists many partially overlapping types
(e.g.\ ``human'', ``theoretical physicist'', ``Nobel laureate'' for
\texttt{Q937}); a single deterministic anchor gives downstream
type-aware prompts a stable, reproducible signal.  The result is
stored as \texttt{type\_triples} together with two lookup
dictionaries (label$\rightarrow$type and QID$\rightarrow$type) inside
the saved graph object.  \texttt{LocalSubgraphClient} merges these
type triples into the same adjacency it builds from
\texttt{candidate\_paths}, so every KG-aware baseline observes
\textit{instance of} and \textit{subclass of} as ordinary outgoing
relations, preserving fairness across methods.

\paragraph{Resulting scale.}
The union of per-question subgraphs across \bench{} covers
approximately 68K Wikidata entities.

\section{Answer Choices Construction}
\label{app:distractors}

Each question in \bench{} is paired with five answer choices
$\mathcal{A}=\{a,b,c,d,e\}$ where option \texttt{e} is fixed to the
string ``None''.  The remaining four options $\{a,b,c,d\}$ are
constructed from the question's gold reasoning path
$p^{*}\!=\!e_0\!\xrightarrow{r_1}\!e_1\!\xrightarrow{r_2}\!\cdots\!\xrightarrow{r_N}\!e_N$
(extracted from Wikidata, $N\!\in\![2,6]$) under one of two regimes
depending on the surface form of the gold answer.  Throughout, the
\emph{answer entity} $e^{*}$ is one of the intermediate nodes
$e_1,\dots,e_{N-1}$ — never a topic entity, since
$e_0$ and $e_N$ are exposed in the question itself.

\paragraph{Regime~1 — Single-entity gold answer.}
When the gold answer is a single entity (e.g., the answer slot reads
just ``\textit{human}''), distractors are drawn directly from the
local Wikidata neighborhood:
\begin{enumerate}
  \item[(i)] Collect $\mathcal{N}_{\le 3}(e^{*})$, the set of entities
    reachable from $e^{*}$ within $3$ hops in Wikidata.
  \item[(ii)] Filter to the type-matched candidates
    $\mathcal{C}\!=\!\{e\in\mathcal{N}_{\le 3}(e^{*})\,:\,\tau(e)=\tau(e^{*})\}$,
    where $\tau$ is the Wikidata \texttt{P31}/\texttt{P279} type
    lookup.  For instance, if $e^{*}$ is a drug, $\mathcal{C}$ contains
    only other drugs.
  \item[(iii)] Sample three entities from $\mathcal{C}$, excluding
    $e^{*}$ and any entity that lies on a path semantically equivalent
    to $p^{*}$.  These three, together with $e^{*}$, form the four
    options $\{a,b,c,d\}$ in random order.
\end{enumerate}
This produces options that are (a) syntactically interchangeable with
the gold answer (same Wikidata type), (b) topologically nearby in the
KG (within 3 hops) and therefore semantically plausible, yet (c)
provably wrong because they do not satisfy the relation chain
prescribed by $p^{*}$.

\paragraph{Regime~2 — Sentence-form gold answer.}
When the gold answer is a full sentence that embeds the answer entity
in natural-language scaffolding (e.g., ``\textit{Liberation has an
effect that produces liberty, which is categorized under a Wikimedia
category that is a subclass of a list.}''), entity-only swapping does
not yield grammatical distractors.  We instead use the following
two-step procedure:
\begin{enumerate}
  \item[(i)] \emph{Anchor extraction.}  We identify the single anchor
    entity $e^{*}$ inside the gold sentence — the entity whose
    replacement turns the sentence into a wrong claim while the rest
    of the wording remains valid.  For the example above the anchor
    is \textit{liberty}.
  \item[(ii)] \emph{Type-matched candidate pool.}  As in Regime~1, we
    collect $\mathcal{C}\!=\!\{e\in\mathcal{N}_{\le 3}(e^{*}):\tau(e)=\tau(e^{*})\}$.
  \item[(iii)] \emph{LLM-based sentence rewriting.}  We provide the
    question, the gold sentence (with $e^{*}$ marked), and the
    candidate pool $\mathcal{C}$ to GPT-5 and prompt it to produce
    three distractor sentences, each constructed by substituting a
    different candidate from $\mathcal{C}$ into the anchor slot and
    making minimal local edits so that the resulting sentence remains
    grammatical and stylistically aligned with the gold sentence. 
\end{enumerate}

\paragraph{Option \texttt{e} is always 'None'}
In a controlled fraction of questions the gold answer is \texttt{e}
(``None''), reflecting the realistic case in which a reasoner should
\emph{decline} to commit when no provided option is supported.  These
are constructed by the same two regimes above, except that the
genuine answer $e^{*}$ (or its sentence form) is \emph{not} placed
into any of $\{a,b,c,d\}$.  All four options are filled with
type-matched, locally adjacent distractors from $\mathcal{C}$, none
of which satisfies $p^{*}$.  Option \texttt{e} = ``None'' becomes the
unique correct answer. 

\paragraph{Why type-matched, $\le 3$-hop distractors?}
The two design choices —  same Wikidata type and locality within
$3$ hops — together rule out the most common shortcut a language
model could exploit:
\begin{itemize}
  \item \emph{Type-match} prevents trivial elimination by surface-level
    category mismatch.  If the answer is a drug, a non-drug
    distractor (e.g., a person or a country) would be ruled out
    instantly without any KG reasoning.
  \item \emph{$\le 3$-hop locality} ensures the distractors are
    \emph{plausible} alternatives in the same KG region — the kind of
    entity a model might hallucinate as a near-miss — rather than
    arbitrary unrelated entities, which again would be rejected
    without genuine reasoning.
\end{itemize}
Together, the four options $\{a,b,c,d\}$ are mutually
indistinguishable on type and rough topology; the only feature that
separates the gold answer from the three distractors is whether the
specific relation chain $r_1,\dots,r_N$ actually connects $e_0$ to
that entity and onwards to $e_N$ in Wikidata.  Solving the question
therefore requires the relation-level reasoning the benchmark is
designed to evaluate, not type filtering or surface heuristics.

\paragraph{Quality control.}
All generated questions undergo a two-stage filter.
(1)~\emph{Shortcut removal}: we automatically discard any question
whose gold path admits a $1$-hop semantic equivalent in the
per-question subgraph (e.g., a single Wikidata predicate that already
connects $e_0$ to $e_N$ with the same meaning), since such a question
is trivially solvable without multi-hop reasoning.
(2)~\emph{Manual review}: the remaining questions are inspected by
the authors to verify that distractors are plausible but unambiguously
wrong, that the gold sentence is fluent, and that no two options are
semantically equivalent.  After both filters,
\bench{} contains \textbf{1055} questions across ~20 Wikidata domains.  The empirical distribution
of correct labels over $\{a,b,c,d,e\}$ is on Table~\ref{tab:label_dist}

\begin{table}[t]
\centering
\small
\begin{tabular}{lccccc}
\toprule
\textbf{Label} & a & b & c & d & e \\
\midrule
\textbf{Pct (\%)} & 21.4 & 26.0 & 17.4 & 21.5 & 13.7 \\
\bottomrule
\end{tabular}
\caption{Distribution of correct labels in \bench{} ($N{=}1{,}055$).}
\label{tab:label_dist}
\end{table}

\section{Our Algorithm(\sbd vs. \onec)}
\label{app:algorithms}

The two scheduling variants of Holistic Path Selection(Algorithm ~\ref{alg:onechance},~\ref{alg:stepbydepth}). Both share
the same three-stage backbone (LLM$_1$: schema-level chain selection,
LLM$_2$: grounded path selection without observing $\mathcal{A}$,
LLM$_3$: answer commitment) but differ in how relation chains $\mathcal{R}$
are exposed: \onec{} feeds them all at once and retries on a non-confident
answer ($\hat{a}\!\in\!\{\texttt{e},\texttt{unsure}\}$, where \texttt{e}
is the dataset's explicit ``insufficient information'' option), excluding
previously selected chains; \sbd{} reveals chains depth by depth and, at
each depth, allows up to $T$ attempts on \texttt{unsure} before advancing
to $d{+}1$.  Auxiliary notation:
$\mathrm{Chains}(\mathcal{P}){:=}\{\mathrm{rel}(p):p\!\in\!\mathcal{P}\}$,
$\mathrm{Chains}_d(\mathcal{P}){:=}\{\mathrm{rel}(p)_{1:d}:p\!\in\!\mathcal{P},|\mathrm{rel}(p)|\!\ge\!d\}$,
and $\mathcal{P}_{\mathbf{r}}\rvert_{d}{:=}\{p_{1:d}:p\!\in\!G_q,\mathrm{rel}(p)_{1:d}\!=\!\mathbf{r}\}$.

\begin{algorithm}[t]
\DontPrintSemicolon
\SetAlgoLined
\SetKwInOut{Input}{Input}
\SetKwInOut{Param}{Hyperparam}
\SetKwInOut{Output}{Output}
\caption{\onec{}}
\label{alg:onechance}
\Input{$Q$, choices $\mathcal{A}$, candidate paths $\mathcal{P}$ in $G_q$}
\Param{$T{=}2$, $K_1{=}5$, $K_2{=}5$}
\Output{predicted answer $\hat{a}$}
\BlankLine
$\mathcal{R}\leftarrow\mathrm{Chains}(\mathcal{P})$\tcp*{all relation chains, any depth}
$\mathcal{S}_{\text{prev}}\leftarrow\varnothing$\;
\For{$t=1,\dots,T$}{
  $\mathcal{R}_{\text{avail}}\leftarrow \mathcal{R}\setminus\mathcal{S}_{\text{prev}}$\;
  $\mathcal{R}^{*}\leftarrow\mathrm{LLM}_1\!\bigl(Q,\,\{\sigma_1(\mathbf{r})\}_{\mathbf{r}\in\mathcal{R}_{\text{avail}}};K_1\bigr)$\;
  \lIf{$\mathcal{R}^{*}=\varnothing$}{\textbf{break}}
  $\mathcal{S}_{\text{prev}}\leftarrow\mathcal{S}_{\text{prev}}\cup\mathcal{R}^{*}$\;
  $\mathcal{P}^{*}\leftarrow\bigcup_{\mathbf{r}\in\mathcal{R}^{*}}\mathcal{P}_{\mathbf{r}}$\tcp*{expand to grounded paths}
  \lIf{$\mathcal{P}^{*}=\varnothing$}{\textbf{continue}}
  $\mathcal{P}^{**}\leftarrow\mathrm{LLM}_2\!\bigl(Q,\,\{\sigma_2(p)\}_{p\in\mathcal{P}^{*}};K_2\bigr)$\tcp*{choice-blind}
  \lIf{$\mathcal{P}^{**}=\varnothing$}{\textbf{continue}}
  $\hat{a}\leftarrow\mathrm{LLM}_3\!\bigl(Q,\mathcal{A},\{\sigma_2(p)\}_{p\in\mathcal{P}^{**}}\bigr)$\;
  \uIf{$\hat{a}\in\{a,b,c,d\}$}{\Return $\hat{a}$\tcp*{confident commit}}
  \lElse{\textbf{continue}\tcp*[h]{$\hat{a}\!\in\!\{e,\text{unsure}\}$: retry}}
}
\Return $\hat{a}$ if $\hat{a}\!\in\!\{a,\dots,e\}$ else $e$\tcp*[h]{$T$ exhausted; commit last or default \texttt{e}}
\end{algorithm}

\begin{algorithm}[t]
\DontPrintSemicolon
\SetAlgoLined
\SetKwInOut{Input}{Input}
\SetKwInOut{Param}{Hyperparam}
\SetKwInOut{Output}{Output}
\caption{\sbd{}}
\label{alg:stepbydepth}
\Input{$Q$, choices $\mathcal{A}$, candidate paths $\mathcal{P}$ in $G_q$}
\Param{$T{=}2$, $D{=}6$, $K_1{=}5$, $K_2{=}5$}
\Output{predicted answer $\hat{a}$}
\BlankLine
\For{$d=1,\dots,D$}{
  $\mathcal{R}_d\leftarrow\mathrm{Chains}_d(\mathcal{P})$\tcp*{$d$-hop prefixes only}
  \lIf{$\mathcal{R}_d=\varnothing$}{\textbf{continue}}
  $\mathcal{S}_{\text{prev}}\leftarrow\varnothing$\;
  \For{$t=1,\dots,T$}{
    $\mathcal{R}^{*}_d\leftarrow\mathrm{LLM}_1\!\bigl(Q,\{\sigma_1(\mathbf{r})\}_{\mathbf{r}\in\mathcal{R}_d\setminus\mathcal{S}_{\text{prev}}};K_1\bigr)$\;
    \lIf{$\mathcal{R}^{*}_d=\varnothing$}{\textbf{continue}}
    $\mathcal{S}_{\text{prev}}\leftarrow\mathcal{S}_{\text{prev}}\cup\mathcal{R}^{*}_d$\;
    $\mathcal{P}^{*}_d\leftarrow\bigcup_{\mathbf{r}\in\mathcal{R}^{*}_d}\mathcal{P}_{\mathbf{r}}\rvert_{d}$\tcp*{truncate to depth $d$}
    \lIf{$\mathcal{P}^{*}_d=\varnothing$}{\textbf{continue}}
    $\mathcal{P}^{**}_d\leftarrow\mathrm{LLM}_2\!\bigl(Q,\{\sigma_2(p)\}_{p\in\mathcal{P}^{*}_d};K_2\bigr)$\;
    \lIf{$\mathcal{P}^{**}_d=\varnothing$}{\textbf{continue}}
    $\hat{a}\leftarrow\mathrm{LLM}_3\!\bigl(Q,\mathcal{A},\{\sigma_2(p)\}_{p\in\mathcal{P}^{**}_d}\bigr)$\;
    \uIf{$\hat{a}\in\{a,b,c,d,e\}$}{\Return $\hat{a}$}
    \lElse{\textbf{continue}\tcp*[h]{same-depth retry on \texttt{unsure}}}
  }
  \tcp*[h]{$T$ attempts exhausted at depth $d$: advance to $d{+}1$}
}
\Return $e$\tcp*{exhausted all depths}
\end{algorithm}

\section{Prompts}
\label{app:prompts}

This appendix reproduces the prompts used by the two proposed methods,
\onec and \sbd. Both run an identical 3-step LLM pipeline on the
per-question subgraph: Step~1 selects relation chains at the schema
level, Step~2 selects grounded paths \emph{with answer choices
intentionally hidden} (mirroring the retrieval stage of every baseline
we compare against -- ToG, PoG, FiDeLiS, KARPA), and Step~3 reads the
final shortlist together with the multiple-choice options and emits an
answer letter.

We give a single template per step.  In every box,
\OC{lines highlighted in this color} appear only in \onec,
\SD{lines highlighted in this color} appear only in \sbd, and all
other lines are shared.  Tokens such as \texttt{\{question\}},
\texttt{\{depth\}}, \texttt{\{rel\_text\}} are runtime-substituted
Python f-string placeholders; \texttt{\{header\}} and the optional
\texttt{\{prev\_text\}} blocks are described in
Section~\ref{app:prompts:shared}.

\subsection{Notation legends and shared building blocks}
\label{app:prompts:shared}

The three notation legends below are spliced into Step~1 and
Steps~2--3.  All three are shared by \onec and \sbd.

\medskip
\noindent\textbf{(L1) Default Step~1 legend} -- topic name visible,
downstream nodes shown by type only:
\begin{verbatim}
Notation legend:
- Each candidate path is shown as:
    TOPIC_NAME:[topic_type] -(rel1)- [type1] -(rel2)- [type2] -...- [typeN]
- TOPIC_NAME (before the first colon) is the actual entity name of the
  topic / starting node -- same across all candidate paths.
- [square brackets] denote the TYPE of an entity from the knowledge graph
  (e.g. [Drug], [Disease], [particle accelerator]).  '[unknown]' means
  the type was not recorded for that slot.
- Downstream node NAMES are intentionally hidden at this step; only their
  types are shown.  You should pick paths based on the topic + structural
  type sequence, not on guessing which specific intermediate entity
  appears.
- Text in (parentheses) denotes a RELATION between two adjacent entities.
\end{verbatim}

\noindent\textbf{(L2) Step~1 legend under the \texttt{schema\_only}
ablation} -- topic name also hidden:
\begin{verbatim}
Notation legend:
- Each candidate path is shown as:
    [type0] -(rel1)- [type1] -(rel2)- [type2] -...- [typeN]
- [square brackets] denote the TYPE of an entity from the knowledge graph
  (e.g. [Drug], [Disease]).  '[unknown]' means the type was not recorded.
- All entity names are intentionally hidden at this step (schema-only
  ablation).  Pick paths based purely on the type/relation structure.
- Text in (parentheses) denotes a RELATION between two adjacent entities.
\end{verbatim}

\noindent\textbf{(L3) Steps~2 and 3 grounded-path legend:}
\begin{verbatim}
Notation legend:
- Each path is shown as:
    name:[type] -(rel)- name:[type] -(rel)- ... -(rel)- name:[type]
- 'name' is the actual entity name; '[type]' is its type from the
  knowledge graph (e.g. 'Aspirin:[Drug]').
- When the type was not recorded, the entity name is repeated in the
  brackets (e.g. 'Cereblon:[Cereblon]') -- treat that as 'type unknown'.
- Text in (parentheses) denotes a RELATION between two adjacent entities.
\end{verbatim}

\noindent\textbf{(N1) Anonymized-graph notice} -- prepended verbatim to
Steps~1, 2, and 3 in the \texttt{--anonymized} ablation (identical for
\onec and \sbd):
\begin{verbatim}
Important: All entity names in this knowledge graph have been replaced with
anonymous IDs (e.g. E_1234).  The IDs themselves carry no semantic meaning.
Reason about paths using only:
  - the relations between entities (text in parentheses), and
  - the entity types (text in square brackets).
Do NOT try to guess what entity an ID refers to.
\end{verbatim}

\noindent\textbf{(N2) Additional anonymized clause} -- appended only at
Step~3 immediately after (N1):
\begin{verbatim}
For answer selection specifically:
- If an ID that appears anywhere along a KG path also appears verbatim in
  one of the answer choices, treat that as direct evidence for that
  choice.
- If a path connects the question's IDs through plausible relations to a
  choice's ID, that is sufficient evidence to commit to that choice.
\end{verbatim}

\paragraph{Header variants for Steps~1 and 2.}
The placeholder \texttt{\{header\}} appearing in the boxes below is one
of three single-sentence headers, chosen by call type.  Substitute
\texttt{\{N\}} = batch size, \texttt{\{B\}} = 0-based batch index,
\texttt{\{T\}} = total batches, \texttt{\{K\}} = per-call top-$K$, and
\texttt{<items>} = ``relation paths'' (Step~1, \onec) /
``relation-path prefixes for the CURRENT depth = \{depth\}''
(Step~1, \sbd) / ``grounded paths'' (Step~2, \onec) /
``grounded paths truncated to the CURRENT depth = \{depth\}''
(Step~2, \sbd).

\begin{itemize}\setlength\itemsep{0pt}
\item \emph{Single batch:}
  \texttt{There are \{N\} candidate <items>.~Pick up to \{K\} most
  promising/informative ones.}
\item \emph{Multi-batch:}
  \texttt{You are reviewing batch \{B+1\} of \{T\} of candidate
  <items>.~This batch contains \{N\} <items>.~After all batches are
  reviewed, your selections will be merged.~Pick up to \{K\} from THIS
  BATCH ONLY.}
\item \emph{Consolidation:}
  \texttt{You previously reviewed multiple batches of candidate <items>
  and short-listed the most promising/informative ones from each
  batch.~Now from the \{N\} finalist <items> below, pick the best up to
  \{K\} for answering the question.}
\end{itemize}

\subsection{Step 1 prompt}
\label{app:prompts:step1}

The two methods diverge on (i)~\onec's \OC{retry banner}, fired only
when \texttt{attempt\,>\,1}, plus its terser closing \OC{\texttt{Rules}}
list, vs.\ (ii)~\sbd's \SD{\texttt{Current search depth} block}, the
\SD{prefix-aware phrasing}, and the longer closing
\SD{\texttt{Important} list} that explicitly permits returning
\texttt{none} so that the depth loop can move on.

\begin{tcolorbox}[promptbox]
\begin{lstlisting}[style=promptstyle]
You are solving a knowledge graph reasoning problem.

[IF anonymized: paste notice (N1) here]

Question:
{question}

(*\OC{=== \onec only -- IF attempt > 1: ===}*)
[RETRY ATTEMPT {attempt}/{max_attempts}] The previous attempt's paths
did not lead to a confident answer.  Pick a DIFFERENT set of relation
paths that explore other parts of the knowledge graph.
(*\OC{=== end \onec block ===}*)

(*\SD{=== \sbd only: ===}*)
Current search depth:
- current depth = {depth}
- deeper rounds may still happen up to depth = {max_depth}
(*\SD{=== end \sbd block ===}*)

{header}

(*\OC{[\onec heading]}*) Candidate relation paths:
(*\SD{[\sbd heading]}*) Candidate relation-path prefixes:
{rel_text}

[paste legend (L1) -- default, OR (L2) under --schema_only ablation]
{prev_text}

(*\OC{=== \onec rule list: ===}*)
Rules:
- The paths may have different hop lengths.
- Prefer semantically relevant paths.
- Do NOT repeat previously selected paths.
- Return ONLY the path numbers, separated by commas.
- Example output: 1,3,5
- If only 1 or 2 paths seem promising, return fewer.
- Do not explain your choice.
(*\OC{=== end \onec rule list ===}*)

(*\SD{=== \sbd rule list: ===}*)
Important:
- The current depth may still be too shallow.
- If the current-depth evidence seems incomplete, that is okay: deeper
  rounds may reveal additional evidence later.
- So do NOT force a path just because you must pick something.  Pick
  only genuinely promising prefixes for the current depth.
- Prefer unexplored paths.
- Do NOT repeat previously selected paths.
- Return ONLY the path numbers, separated by commas.
- Example output: 1,3,5
- If only 1 or 2 paths look promising, return fewer.
- If none look useful, return: none
- Do not explain your choice.
(*\SD{=== end \sbd rule list ===}*)
\end{lstlisting}
\end{tcolorbox}

\noindent The optional \texttt{\{prev\_text\}} block listing
already-selected paths to be avoided takes one of two forms.
\OC{In \onec:} \texttt{"Previously selected paths from earlier attempts
(DO NOT repeat and DO NOT pick paths that share the same relation prefix
unless you have a strong new reason -- earlier attempts already explored
them):"} followed by a bulleted list of typed chains.
\SD{In \sbd:} \texttt{"Previously selected paths at this depth (DO NOT
repeat):"} followed by a bulleted list of typed chains.

\subsection{Step 2 prompt}
\label{app:prompts:step2}

\sbd reuses the same \SD{depth block} as in Step~1 and inserts a single
phrase ``\SD{at the current depth}'' inside the otherwise-shared rule
list; \onec omits both.  Crucially, \emph{no answer choices} are
exposed at this step in either method.

\begin{tcolorbox}[promptbox]
\begin{lstlisting}[style=promptstyle]
You are solving a knowledge graph reasoning problem.

[IF anonymized: paste notice (N1) here]

Question:
{question}

(*\SD{=== \sbd only: ===}*)
Current search depth:
- current depth = {depth}
- deeper rounds may still happen up to depth = {max_depth}
(*\SD{=== end \sbd block ===}*)

{header}

Candidate paths:
{rel_text}

[paste legend (L3) -- grounded variant]

Rules:
- Pick the paths whose entities and relations most directly bear on the
  question(*\SD{ at the current depth}*).
- The answer entity may appear at any position along a path (intermediate
  or end).
- Prefer paths that are concrete and specific over generic structural
  matches.
- Do NOT consider any answer choices -- at this step you have not been
  shown any.  Focus only on which paths best ground the question.
- Return ONLY the path numbers, separated by commas.
- Example output: 1,3,5
- If only 1 or 2 paths seem informative, return fewer.
- Do not explain your choice.
\end{lstlisting}
\end{tcolorbox}

\subsection{Step 3 prompt}
\label{app:prompts:step3}

The two methods diverge most clearly here.  \onec \emph{forces} a commit
to one of $\{a,b,c,d,e\}$ -- there is no abstention option -- whereas
\sbd explicitly permits the literal token \texttt{unsure} so that the
outer loop can advance to a deeper round; \sbd also reuses the depth
block from Steps~1--2.

\begin{tcolorbox}[promptbox]
\begin{lstlisting}[style=promptstyle]
[IF anonymized: paste notice (N1), then notice (N2)]

Question:
{question}

Choices:
a. {choices['a']}
b. {choices['b']}
c. {choices['c']}
d. {choices['d']}
e. {choices['e']}

(*\SD{=== \sbd only: ===}*)
Current search depth:
- current depth = {depth}
- deeper rounds may still happen up to depth = {max_depth}
(*\SD{=== end \sbd block ===}*)

[paste legend (L3) -- grounded variant]

(*\OC{=== \onec evidence heading: ===}*)
Knowledge graph paths (the answer entity may appear at any position
along a path, including intermediate nodes -- not only at the end):
(*\OC{=== end \onec heading ===}*)
(*\SD{=== \sbd evidence heading: ===}*)
Current evidence from knowledge graph paths
(the answer entity may appear at any position along a path, including
intermediate nodes -- not only at the end):
(*\SD{=== end \sbd heading ===}*)
{path_text}

(*\OC{=== \onec decision rule: ===}*)
Decision rule:
- You MUST commit to exactly one option among a / b / c / d / e.
- Pick the choice whose entity (or its synonym/parent class) best matches
  the entities and types observed along the paths above.
- The answer entity may be at an intermediate hop; the path does not need
  to end at the answer.
- Do NOT explain.  Return only the option letter (a, b, c, d, or e).
(*\OC{=== end \onec decision rule ===}*)

(*\SD{=== \sbd decision rule: ===}*)
Decision rule:
- If the CURRENT evidence is already sufficient to justify one answer
  strongly, choose one option (a/b/c/d/e).
- If the CURRENT evidence is incomplete, ambiguous, or you think a
  deeper search could still change the answer, respond with: unsure
- Do NOT guess just because an answer choice looks plausible.
- Be conservative: deeper depths may reveal missing links.

Return format:
- Return the option letter first if you commit to an answer.
- Otherwise return exactly: unsure
(*\SD{=== end \sbd decision rule ===}*)
\end{lstlisting}
\end{tcolorbox}

\section{Models and serving.}
\label{app:resource}
We evaluate all methods with two backbone LLMs.
\textbf{Qwen3-32B} (\texttt{Qwen/Qwen3-32B}) is served locally in
BF16 precision via vLLM with tensor parallelism across
$4\times$NVIDIA A100-SXM4-80GB GPUs (CUDA~12.2,
\texttt{tensor\_parallel\_size{=}4}), exposed through an
OpenAI-compatible HTTP endpoint.
\textbf{GPT-5-mini} (\texttt{gpt-5-mini-2025-08-07}) is accessed
through the OpenAI API with \texttt{max\_tokens{=}128{,}000},
\texttt{reasoning\_effort{=}minimal}, and \texttt{verbosity{=}low}.
For the small number of components that require text embeddings
(KAPING and FiDeLiS retrieval, MCQ-letter parsing fallback), all
methods uniformly use OpenAI \texttt{text-embedding-3-small},
independent of the backbone LLM.  Routing both Qwen3-32B and
GPT-5-mini through the same OpenAI-compatible interface ensures
that every baseline and our method share an identical inference
contract — only the underlying model weights differ.

\section{Question Generation}
\label{app:questionGen}

\begin{tcolorbox}[
  enhanced,
  breakable,
  colback=gray!3, colframe=gray!50!black,
  title={\textbf{Prompt: Selecting Good Paths and Writing Compressed Questions}},
  fonttitle=\bfseries\small,
  coltitle=white, colbacktitle=gray!60!black,
  arc=2pt, boxrule=0.5pt,
  left=10pt, right=10pt, top=8pt, bottom=8pt,
  fontupper=\footnotesize,
]

\textbf{Task overview.}
You are building a Knowledge-Graph-based Question Answering dataset.
Each question must sound simple and natural, but require reasoning
over multiple hidden relations in the KG to find the answer. Your
job is to select paths that can be meaningfully compressed and to
write compressed questions.

\bigskip
\textbf{$\checkmark$ Good path / well-compressed question.}
A good (path, question) pair satisfies most of the following.

\begin{enumerate}
  \item \textbf{Hidden multi-hop reasoning.}\;
        The question gives no explicit clue about how many hops or
        what relations are involved. A reader cannot reconstruct
        the underlying path structure from the question alone.
        \\[2pt]
        \emph{Example.}\;
        Q: ``If someone has a BRCA1 mutation, who is more at risk?''
        \;\;Path: BRCA1 $\to$ mutation can cause $\to$ breast cancer
        $\to$ related to $\to$ female health.

  \item \textbf{Semantic compression.}\;
        The question summarises the overall semantics of the full
        path in a concise, natural way. Intermediate concepts and
        entities are not mentioned; the reasoning is embedded
        implicitly through conceptual verbs such as \emph{affect},
        \emph{relate to}, \emph{lead to}, \emph{enable}, or
        \emph{associated with}.

  \item \textbf{Natural, human-like curiosity.}\;
        The question reads like something a human would ask in
        conversation or in everyday science reasoning---not like a
        SPARQL or database query. It is short, smooth, and does not
        list relation names.

  \item \textbf{Functional or causal abstraction.}\;
        The question focuses on effects, roles, or relationships
        rather than structural links.
        \\[2pt]
        \emph{Examples.}\;
        ``What symptom does this drug help with?'' hides a
        biochemical chain (drug--gene--protein action).
        ``Who is Jack's cousin?'' hides the parent--sibling--child
        path.

  \item \textbf{Non-trivial reasoning needed.}\;
        The answer can only be reached by combining multiple
        knowledge-graph edges; no single triple in the KG directly
        connects the start entity to the answer entity.

  \item \textbf{Entity-level coherence.}\;
        The start entity and the answer entity have an interpretable
        real-world connection (drug~$\leftrightarrow$~symptom,
        gene~$\leftrightarrow$~sex, person~$\leftrightarrow$~relative).
        The reasoning path between them expresses a plausible
        causal, hierarchical, or kinship relation.
\end{enumerate}

\bigskip
\textbf{$\times$ Bad path / poorly-compressed question.}
Avoid paths that exhibit one or more of the following flaws.

\begin{enumerate}
  \item \textbf{Explicit relation leakage.}\;
        The question directly repeats or paraphrases KG predicates.
        \\[2pt]
        \emph{Bad example.}\;
        ``Which gene encodes the target that serotonin binds to?''
        \;\;Path: serotonin $\to$ interacts with $\to$ receptor
        $\to$ encoded by $\to$ gene.
        The relations ``binds to'' and ``encodes'' are exposed
        verbatim.

  \item \textbf{Predictable reasoning chain.}\;
        From the question, it is clear that two or three hops are
        needed. If a reader can guess the intermediate node types,
        the path is not compressed.

  \item \textbf{Over-specific or mechanical wording.}\;
        Uses technical or structural phrasing like ``the $X$ that
        $Y$ binds to'' or ``entity~1 encoded by entity~2''. Sounds
        like a database query rather than a human question.

  \item \textbf{Irrelevant or weak semantic chain.}\;
        The path connects unrelated domains, or yields an answer
        that does not make conceptual sense together with the start
        entity.
\end{enumerate}

\bigskip
\textbf{Output format.} Respond strictly in the following JSON
schema:

\begin{verbatim}
{
  "judge"   : "QA generation appropriate"
              | "QA generation inappropriate",
  "question": "<string, only if appropriate>" | "None",
  "answer"  : "<string, only if appropriate>" | "None",
  "reason"  : "<brief reason when inappropriate>" | "None"
}
\end{verbatim}

\bigskip
\textbf{Examples.}

\smallskip
\textit{[Example 1]}
\begin{verbatim}
KG path: Jack -- Father is -- Sibling is -- Child is -- Sujan
relation meaning: Father is = male parent;
                  Sibling is = individuals sharing at least one parent;
                  Child is = ...
judge   : "QA generation appropriate"
question: "Who is Jack's cousin?"
answer  : "Sujan"
\end{verbatim}

\smallskip
\textit{[Example 4]}
\begin{verbatim}
KG path: Albert Einstein -- influenced by -- influenced by
         -- field of work -- physics
relation meaning: influenced by = was inspired by;
                  field of work = specialization of a person/org
judge   : "QA generation inappropriate"
question: "None"
answer  : "None"
reason  : "too broad and generic"
\end{verbatim}

\bigskip
\textit{Now, it's your turn.}
\begin{verbatim}
KG path: [Candidate Paths]
relation meaning: [Meaning of each relation from Wikidata]
judge   :
question:
answer  :
\end{verbatim}

\end{tcolorbox}
\captionof{figure}{Prompt used to filter Wikidata paths for
compressibility and to generate the natural-language questions
that anchor each \bench{} item. Path selection is performed in a
single GPT-5-mini call per candidate path; only paths satisfying
the criteria above are retained.}
\label{fig:prompt_compression}

\section{Case Study}
\label{app:qualtative_anal}
We present three log-grounded case studies, one per existing KGQA paradigm, in which a representative method fails on \bench\ while our method succeeds.

\subsection{KAPING vs.\onec}
We trace KAPING~\citep{kaping} and \onec\ (\methodname) on the same
\emph{compressed-only} item from \bench\  to
illustrate the failure mode argued in \S\ref{sec:intro:hidden}
quantitatively. Both systems are given the identical question, topic
entity, and underlying KG; they differ only in their reasoning
paradigm.

\paragraph{Question and gold KG path.}
\begin{quote}
\textit{``Which inherited disorder might Luise von der Pfalz be at
elevated risk for because of family history?''}\\
Options: (a)~scarlet fever\quad
\textbf{(b)~gout}\quad
(c)~leprosy\quad
(d)~cystitis\quad
(e)~None
\end{quote}
The question surface contains only two content phrases, \emph{inherited
disorder} and \emph{family history}, neither of which lexically
matches any relation on the 6-hop gold path:
\begin{quote}\small
Luise von der Pfalz $\xrightarrow{\text{father}}$ Charles I Louis,
Elector Palatine $\xrightarrow{\text{mother}}$ Elizabeth Stuart, Queen
of Bohemia $\xrightarrow{\text{father}}$ James VI and I
$\xrightarrow{\text{medical condition}}$ \textbf{gout}
$\xrightarrow{\text{subclass of}}$ genetic disease.
\end{quote}

\paragraph{KAPING trace (predicts \emph{e.~None}, incorrect).}
KAPING retrieves the top-$10$(default hyper-parameter in paper) triples around the topic entity by
question--triple embedding similarity (Table~\ref{tab:kaping_qid500}).
Two observations explain the failure. First, three of the six retrieved
triples are surface-overlap noise---triples whose subject or object
shares the entity name string ``Luise''---rather than evidence about
the question. Second, only the first two hops of the gold chain
(\emph{father}, \emph{mother}) are recovered; the remaining four hops,
including the triple containing the answer entity \emph{gout}, are
absent from the retrieved set. With no evidence supporting any
specific disorder, the LLM defaults to the safe option \emph{e.~None}.

\begin{table}[h]
\centering
\small
\begin{tabular}{c l l l c}
\toprule
\# & Subject & Relation & Object & Role \\
\midrule
1 & Luise von der Pfalz & given name        & Luise               & noise \\
2 & Luise von der Pfalz & \textbf{father}   & Charles I Louis     & gold hop 1 \\
3 & Luise               & different from    & Luise               & noise \\
4 & Charles I Louis     & \textbf{mother}   & Elizabeth Stuart    & gold hop 2 \\
5 & Luise               & writing system    & Latin script        & noise \\
6 & Charles I Louis     & owner of          & Portrait of a Man\ldots & noise \\
7 & ...\\
\bottomrule
\end{tabular}
\caption{KAPING top-6 retrieval. Three of six
slots are consumed by entity-name surface-overlap distractors; gold
hops 3--6, including the answer entity \emph{gout}, are not retrieved.}
\label{tab:kaping_qid500}
\end{table}

\paragraph{\onec\ trace (predicts \emph{b.~gout}, correct).}
\onec\ enumerates all 6-hop relation chains around the topic entity
(958 candidate paths, 904 unique relation sequences) and exposes them
to the LLM in a single view. The model selects three indices in one
step; all three share the same first five relations as the gold path
(Table~\ref{tab:onec_qid500}). Grounding these chains to the KG yields
a single entity-instantiated path
\begin{quote}\small
Luise $\to$ Charles I Louis $\to$ Elizabeth Stuart $\to$ James VI and I
$\to$ \textbf{gout} $\to$ genetic disease $\to \cdots$,
\end{quote}
which makes the answer entity \emph{gout} explicitly visible in the
final reasoning input. The LLM then selects \emph{b.~gout}.

\begin{table}[h]
\centering
\small
\begin{tabular}{c l}
\toprule
\# & Selected relation chain (depth 6) \\
\midrule
1 & father $\to$ mother $\to$ father $\to$ medical condition $\to$ subclass of $\to$ \emph{age of onset} \\
2 & father $\to$ mother $\to$ father $\to$ medical condition $\to$ subclass of $\to$ \emph{has cause} \\
3 & father $\to$ mother $\to$ father $\to$ medical condition $\to$ \emph{health specialty} $\to$ subclass of \\
\bottomrule
\end{tabular}
\caption{Relation chains selected by \onec\ 
in a single step. All three share the gold 5-hop prefix
(\emph{father, mother, father, medical condition, subclass of}), which
no relation alone matches lexically with the question phrase
``family history''---the alignment is recoverable only at the
\emph{chain} level.}
\label{tab:onec_qid500}
\end{table}

\paragraph{Discussion.}
The two traces concretely instantiate the alignment-collapse mechanism
of \S\ref{sec:intro:hidden}. KAPING's failure is not an accuracy
fluctuation but a structural consequence of similarity-based
retrieval under compression: with no surface signal aligning
\emph{family history} to \emph{father}/\emph{mother}/\emph{medical
condition}/\emph{subclass of}, ranking is dominated by entity-name
string overlap, the retrieval budget is exhausted by noise, and the
gold answer entity never enters the LLM's evidence window. \onec\
succeeds for the symmetric reason: by exposing all candidate chains
jointly, it lets the LLM align the \emph{whole} chain
\emph{father$\to$mother$\to$father$\to$medical condition$\to$subclass
of} with \emph{family history} as a single semantic unit, even though
no individual relation on the chain matches the question's surface.
This is precisely the regime in which hop-local cues are
uninformative and only chain-level semantics carries signal.

\subsection{KARPA vs.\ \sbd}
\label{appendix:qualitative_qid158}

We trace KARPA~\citep{karpa} and \sbd\ (\methodname) on the same
\emph{cause-and-effect} item from \bench\ to
illustrate the failure mode argued in \S\ref{sec:intro:hidden}: even
when the gold relation is present in the candidate set, the
plan-then-retrieve LLM commits to a different relation, and the
failure cascades through the remaining stages.

\paragraph{Question and gold KG path.}
\begin{quote}
\textit{``Through which economic sector does habitat fragmentation
most plausibly relate to the economy?''}\\
Options: (a)~\emph{chemical industry}\quad
(b)~\emph{caretaking}\quad
\textbf{(c)~\emph{agriculture}}\quad
(d)~\emph{restoration}\quad
(e)~None
\end{quote}
\begin{quote}\small
habitat fragmentation $\xrightarrow{\text{has cause}}$
\textbf{agriculture} $\xrightarrow{\text{instance of}}$ economic
sector $\xrightarrow{\text{part of}}$ economy.
\end{quote}
The question surface contains \emph{economic sector} and
\emph{economy} but no token aligned with the gold first-hop relation
\emph{has cause}.

\paragraph{KARPA trace (predicts \emph{d.~restoration}, incorrect).}
KARPA proceeds through six stages
(Table~\ref{tab:karpa_qid158}). At \texttt{relation\_extraction},
\emph{both} \emph{has cause} and \emph{has effect} are retained
among the 25 candidate relations; this isolates the failure from any
retrieval-recall issue. The subsequent \texttt{re\_planning} stage
nonetheless commits to the plan
\emph{relates to sustainable development goal, target or indicator}
$\to$\emph{industry} (length-2), with a length-3 variant
\emph{has effect}$\to$\emph{relates to SDG...}$\to$\emph{industry}.
Neither plan contains \emph{has cause}. The embedding-based
\texttt{matching} stage then grounds these plans into eight chains
(Table~\ref{tab:karpa_matching_qid158}): the six top-scored chains
are 3- to 6-hop paths through SDG entities such as \emph{Target~6.6}
and \emph{SDG~6}, while two short \emph{has effect}-based chains
ending in \emph{urbanization} and \emph{Ecosystem decay} appear at
the bottom with the lowest scores. No chain in the retrieved set
uses \emph{has cause}. At \texttt{reasoning}, the LLM dismisses the
top six SDG chains as not naming an economic sector, latches onto
path~7 (\emph{has effect}$\to$\emph{has contributing factor},
\emph{urbanization}), and emits the free-form answer
\emph{\{construction and services sectors\}}---an inference about
urbanization that is not among the five options. The
\texttt{mcq\_mapping} stage then resolves this off-list answer to
option~\emph{d.\ restoration}.

\begin{table}[h]
\centering
\small
\begin{tabular}{l p{8.4cm}}
\toprule
Stage & Output (verbatim from log, abridged) \\
\midrule
\texttt{initial\_planning}     & \texttt{flat\_relations}: \emph{environmental\_impact}, \emph{economic\_sector} (free-form tokens not in KG schema). \\
\texttt{relation\_extraction}  & 25 KG relations retained; \emph{both} \emph{has cause} (gold) \emph{and} \emph{has effect} included. \\
\texttt{re\_planning}          & Length-2 plan: \emph{relates to SDG, target or indicator}$\to$\emph{industry}. Length-3 variant prepends \emph{has effect}. Neither plan uses \emph{has cause}. \\
\texttt{matching}              & 8 grounded chains; see Table~\ref{tab:karpa_matching_qid158}. None uses \emph{has cause}. \\
\texttt{reasoning}             & Selects path 7 (\emph{has effect $\to$ urbanization}); emits \emph{\{construction and services sectors\}} (not in option list). \\
\texttt{mcq\_mapping}          & Resolves off-list answer to \emph{d.\ restoration}. \\
\bottomrule
\end{tabular}
\caption{KARPA pipeline trace. Although the
gold relation \emph{has cause} is retained at
\texttt{relation\_extraction}, \texttt{re\_planning} commits to a
plan that does not include it, and the failure cascades through
matching, reasoning, and option mapping.}
\label{tab:karpa_qid158}
\end{table}

\begin{table}[h]
\centering
\small
\begin{tabular}{c c l}
\toprule
\# & Score & Grounded chain (verbatim, head $\to$ tail entity) \\
\midrule
1 & 0.631 & h.f.\ $\to$ class of object(s) of occurrence $\to$ different from $\to$ relates to SDG, \emph{Target 6.6} \\
2 & 0.627 & \dots $\to$ relates to SDG $\to$ facet of, \emph{SDG 6} \\
3 & 0.614 & \dots $\to$ relates to SDG $\to$ instance of, \emph{SDG Target} \\
4 & 0.595 & \dots $\to$ facet of $\to$ instance of, \emph{SDG} \\
5 & 0.563 & \dots $\to$ instance of $\to$ subclass of, \emph{goal} \\
6 & 0.561 & \dots $\to$ instance of $\to$ subclass of $\to$ subclass of, \emph{condition} \\
\midrule
7 & 0.451 & h.f.\ $\to$ has effect $\to$ has contributing factor, \emph{urbanization} \\
8 & 0.444 & h.f.\ $\to$ has effect, \emph{Ecosystem decay} \\
\bottomrule
\end{tabular}
\caption{KARPA \texttt{matching} top-8 grounded chains. (h.f.\ = habitat fragmentation). The six
top-scored chains all enter SDG-related sub-graphs through
\emph{class of object(s) of occurrence $\to$ different from}; chains
7--8 use \emph{has effect} but score lower. \emph{Has cause}
(gold first-hop) does not appear in any chain because the
\texttt{re\_planning} step did not include it.}
\label{tab:karpa_matching_qid158}
\end{table}

\paragraph{\sbd\ trace (predicts \emph{c.\ agriculture}, correct).}
\sbd\ enumerates 2{,}041 candidate paths up to depth~6 and grows the
visible chain set one depth at a time .
At depth~1 the LLM sees seven unique relation paths and flags
\emph{has cause} as the most promising frontier relation, but issues
\texttt{unsure} since option~\emph{c} requires a chain that reaches
\emph{economy}. At depth~3 the visible chain extends to
\emph{habitat fragmentation $\to$ (has cause) $\to$ agriculture
$\to$ (instance of) $\to$ economic sector $\to$ (part of) $\to$
economy}, which exactly matches option~\emph{c}; the LLM commits at
depth~3 (step~3 of the depth schedule).

\paragraph{Discussion.}
The traces isolate a specific failure mode of plan-then-retrieve:
the gold relation \emph{has cause} was visible to KARPA at
\texttt{relation\_extraction} but was discarded by the LLM's own
planning step in favour of \emph{relates to sustainable development
goal, target or indicator}. Once the plan is fixed, the matching
stage---constrained to ground that plan---retrieves only chains
using the planned relations; the gold chain never enters the search
space. The reasoning step, working with the resulting eight
SDG-or-\emph{has effect} chains, can no longer reach
\emph{agriculture} and produces an answer (\emph{construction and
services sectors}) that lies outside the option set, forcing an
arbitrary final mapping. \sbd\ avoids the cascade by removing the
planning step altogether: it presents the KG-grounded relation set
directly, and the LLM's only decision is which relations to
instantiate. The case suggests that the LLM-mediated planning step
in plan-then-retrieve is not merely a soft prefilter but a hard
commitment point that, once mis-set, cannot be recovered downstream.

\subsection{ToG vs. \methodname}
\begin{figure*}[t]
\centering
\includegraphics[width=\textwidth]{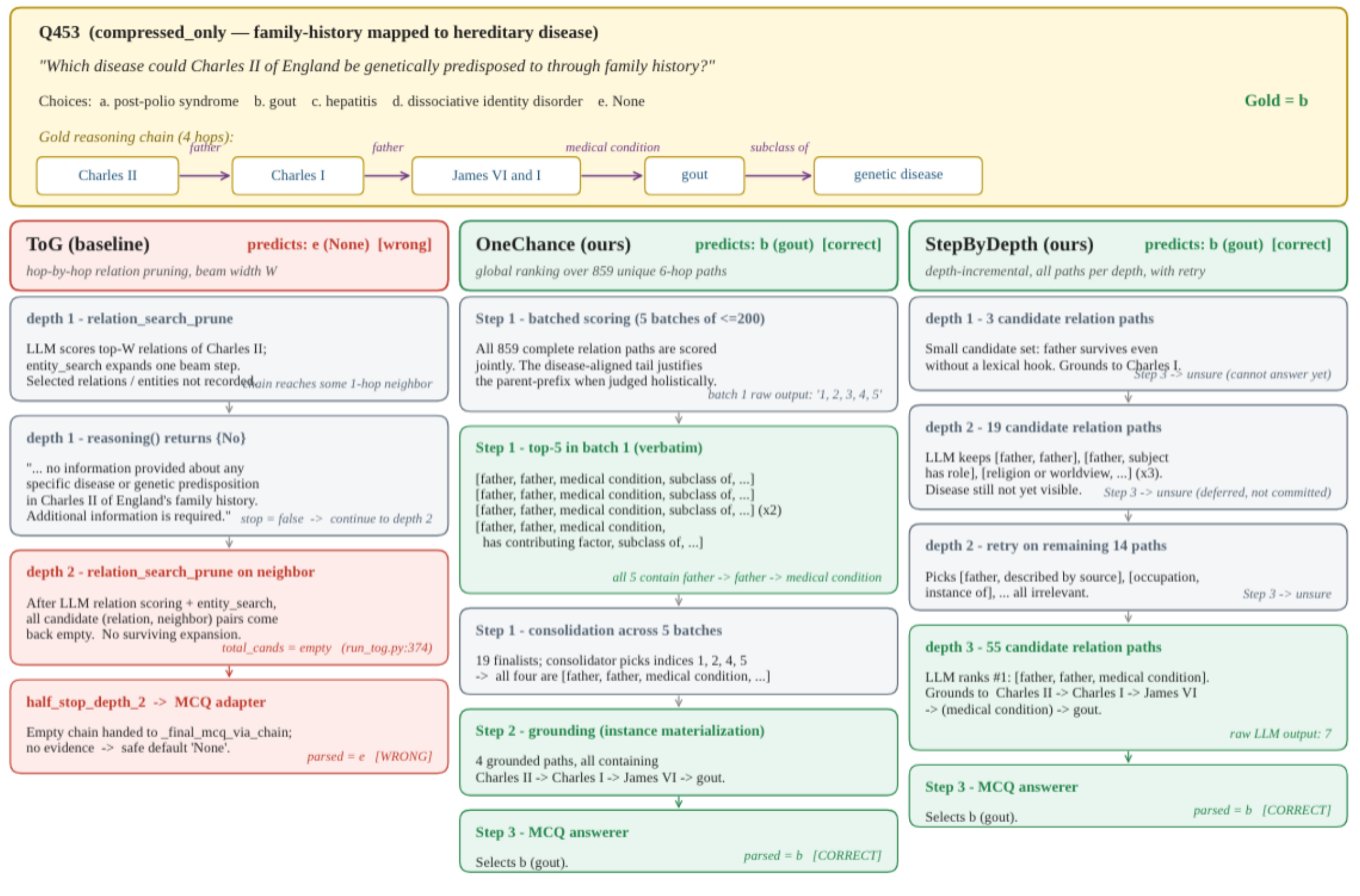}
\caption{\textbf{Trace comparison , a representative
\emph{compressed\_only} question.} The surface phrase ``family history''
silently compresses two \emph{father} hops; only the path's tail
(\emph{medical condition}) carries a lexical anchor in the question
(top, gold chain). Each panel below reproduces one method's trace
from the log.
\textbf{ToG} reaches a 1-hop neighbor at depth~1; the LLM judge
correctly defers with \texttt{\{No\}}. At depth~2, after relation
scoring and \texttt{entity\_search}, the candidate beam is empty
; \texttt{half\_stop} fires and the
empty-chain MCQ adapter falls back to ``None''. The specific
relations selected by the depth-2 LLM scorer are not recorded in the
existing trace; the failure is observable only as the empty
intersection of relation pruning and entity expansion. 
\textbf{OneChance} ranks all 859 length-6 relation paths jointly;
in the first batch, the top-5 LLM-selected paths are \emph{all} variants
of \texttt{[father, father, medical condition, subclass of, \dots]},
despite \emph{father} having no surface counterpart in the question.
\textbf{StepByDepth} keeps \emph{father} at depth~1 from a 3-path
candidate set, defers via \emph{unsure} while the disease tail is not
yet visible, and at depth~3 ranks \texttt{[father, father, medical
condition]} \#1 of 55 candidates, grounding to \emph{gout}.
The contrast isolates the failure mode discussed in
\S\ref{res:hopbypruning_prob}: locally-scored, hop-by-hop pruning is
non-recoverable when path--question alignment is concentrated at the
path's tail, while methods that defer commitment until later (per-depth
\emph{unsure}) or until the entire path is visible (global path
ranking) survive the same alignment gap.}
\label{fig:q453_case_study}
\end{figure*}

Figure~\ref{fig:q453_case_study} shows the selection trajectories by which ToG arrives at the wrong answer while \sbd\ and \onec\ arrive at the correct one.

\section{Generalization of Holistic Path Selection}
\label{app:generalization}

To test whether the advantage of Holistic Path Selection is specific to \bench or Wikidata, we evaluate it on transformed samples from two additional KGQA benchmarks: MetaQA, built on WikiMovies, and ComplexWebQuestions (CWQ), built on Freebase. For each benchmark, we preserve the original entities, answers, and gold paths while weakening explicit lexical cues that reveal the underlying relation sequence.

\subsection{MetaQA}

We transform 100 two- and three-hop MetaQA questions by concealing relation-level lexical cues, excluding one-hop shortcuts, and preserving semantic alignment with the original gold path.

\noindent\textbf{Example.}
The path

\[
\begin{gathered}
\textit{A Bridge Too Far}
\xrightarrow{\textit{directed by}}
\textit{Richard Attenborough}
\xleftarrow{\textit{directed by}}
\textit{Closing the Ring}
\\
\textit{Closing the Ring}
\xrightarrow{\textit{starred actors}}
\textit{Christopher Plummer}
\end{gathered}
\]

corresponds to the original question, `Who are the actors in the movies directed by the director of \textit{A Bridge Too Far}?'' We transform it into `Who is a leading face in another film by the same creative mind behind \textit{A Bridge Too Far}?'' The transformed question preserves the reasoning path while removing explicit relation cues.

\begin{table}[t]
\centering
\small
\renewcommand{\arraystretch}{1.1}
\begin{tabular*}{\textwidth}{
  @{\extracolsep{\fill}} l *{6}{c} @{}
}
\toprule
Method
& \multicolumn{2}{c}{GPT-5-mini}
& \multicolumn{2}{c}{Claude-Sonnet-5}
& \multicolumn{2}{c}{Gemini-2.5-Pro} \\
\cmidrule(lr){2-3}
\cmidrule(lr){4-5}
\cmidrule(lr){6-7}
& Original & Transformed
& Original & Transformed
& Original & Transformed \\
\midrule
Zero-shot
& 63 & 56
& 77 & 74
& 85 & 80 \\
KARPA
& 70 & 46
& 72 & 52
& 77 & 65 \\
ToG
& 63 & 60
& 89 & 89
& 83 & 83 \\
FiDeLiS
& 83 & 73
& 90 & 87
& 85 & 78 \\
PoG
& 78 & 57
& 80 & 71
& 86 & 69 \\
R2-KG
& 79 & 55
& 90 & 83
& 87 & 83 \\
\midrule
TieredHolistic
& \textbf{94} & \underline{88}
& \underline{97} & \underline{95}
& \underline{98} & \underline{95} \\
FlatHolistic
& \textbf{94} & \textbf{94}
& \textbf{98} & \textbf{98}
& \textbf{100} & \textbf{98} \\
\bottomrule
\end{tabular*}
\caption{Accuracy (\%) on the original and transformed MetaQA questions.
Bold and underline denote the best and second-best results, respectively.}
\label{tab:metaqa_generalization}
\end{table}

Both Holistic variants outperform all baselines across all three LLM backbones. On transformed MetaQA, FlatHolistic achieves $94$--$98\%$ accuracy, compared with $73$--$89\%$ for the strongest baseline in each setting. Transformation reduces FlatHolistic and TieredHolistic by only $0.7$ and $3.7$ points on average, respectively, whereas KARPA and PoG decline by $12$--$24$ and $9$--$21$ points. These results show that Holistic Path Selection remains robust when explicit relation cues are concealed.

\FloatBarrier

\subsection{ComplexWebQuestions}

We additionally transform 100 two-hop CWQ questions by removing or paraphrasing relation-level lexical cues and, where possible, compressing the two-hop relation sequence into a shorter expression.

\noindent\textbf{Example.}
The path

$$
\textit{NYU Violets Men's Basketball}
\xrightarrow{\textit{sports team of}}
\textit{New York University}
\xrightarrow{\textit{founder}}
\textit{Mordecai Manuel Noah}
$$

corresponds to the original question, `Who are the founders of the educational institution where the sports team NYU Violets Men's Basketball plays?'' We transform it into `The institution fielding NYU Violets Men's Basketball was set up by whom?''

\begin{table}[t]
\centering
\small
\renewcommand{\arraystretch}{1.1}
\begin{tabular*}{\textwidth}{
  @{\extracolsep{\fill}} l *{6}{c} @{}
}
\toprule
Method
& \multicolumn{2}{c}{GPT-5-mini}
& \multicolumn{2}{c}{Claude-Sonnet-5}
& \multicolumn{2}{c}{Gemini-2.5-Pro} \\
\cmidrule(lr){2-3}
\cmidrule(lr){4-5}
\cmidrule(lr){6-7}
& Original & Transformed
& Original & Transformed
& Original & Transformed \\
\midrule
Zero-shot
& 70 & 66
& 81 & 77
& 85 & 83 \\
KARPA
& 63 & 59
& 85 & 78
& 81 & 74 \\
ToG
& 61 & 54
& 90 & 82
& 79 & 70 \\
FiDeLiS
& 83 & 71
& 97 & 88
& 80 & 72 \\
PoG
& 83 & 74
& 91 & 82
& 78 & 65 \\
R2-KG
& 78 & 70
& 98 & 89
& 75 & 69 \\
\midrule
TieredHolistic
& \underline{93} & \underline{89}
& \underline{94} & \underline{93}
& \underline{95} & \underline{93} \\
FlatHolistic
& \textbf{96} & \textbf{93}
& \textbf{99} & \textbf{98}
& \textbf{99} & \textbf{98} \\
\bottomrule
\end{tabular*}
\caption{Accuracy (\%) on the original and transformed CWQ questions.
Bold and underline denote the best and second-best results, respectively.}
\label{tab:cwq_generalization}
\end{table}

Both Holistic variants again outperform all baselines across every backbone and setting. KG-based baselines decline by $4$--$13$ points after transformation, whereas TieredHolistic and FlatHolistic decrease by only $2.3$ and $1.7$ points on average. Several KG-based baselines also fall below the closed-book zero-shot model, suggesting that misaligned retrieved paths can introduce distracting context. In contrast, the Holistic variants remain robust to weakened question--path alignment.

The decline on CWQ is smaller than on MetaQA, likely because the sampled CWQ questions contain only two hops and therefore offer fewer opportunities for erroneous hop-by-hop pruning. Nevertheless, MetaQA and CWQ reproduce the same overall pattern across different domains, backbone KGs, and LLM families: weakening lexical alignment disproportionately harms existing path-retrieval methods, while Holistic Path Selection retains its advantage. This supports the generality of both alignment collapse and holistic chain evaluation beyond \bench and Wikidata.

\FloatBarrier


\newpage
\section*{NeurIPS Paper Checklist}

\begin{enumerate}

\item {\bf Claims}
    \item[] Question: Do the main claims made in the abstract and introduction accurately reflect the paper's contributions and scope?
    \item[] Answer: \answerYes{}
    \item[] Justification: The abstract and introduction state the paper's three main contributions: the HiddenPathQA benchmark, the holistic path selection paradigm, and the pseudoword analysis. 

\item {\bf Limitations}
    \item[] Question: Does the paper discuss the limitations of the work performed by the authors?
    \item[] Answer: \answerYes{}
    \item[] Justification: Section Conclusions and Limitations discusses limitations, including reliance on Wikidata coverage.

\item {\bf Theory assumptions and proofs}
    \item[] Question: For each theoretical result, does the paper provide the full set of assumptions and a complete (and correct) proof?
    \item[] Answer:\answerNA{}
    \item[] Justification:The paper does not present formal theoretical results, theorems, or proofs; its contributions are a benchmark, empirical analyses, and evaluation methods.

    \item {\bf Experimental result reproducibility}
    \item[] Question: Does the paper fully disclose all the information needed to reproduce the main experimental results of the paper to the extent that it affects the main claims and/or conclusions of the paper (regardless of whether the code and data are provided or not)?
    \item[] Answer: \answerYes{}
    \item[] Justification: The paper reports the dataset, evaluation metrics, model settings, prompting protocol, and baseline configurations in Evaluation Protocol Sections and Appendices. We also provide data and code repositories with scripts for reproducing the main experiments.

\item {\bf Open access to data and code}
    \item[] Question: Does the paper provide open access to the data and code, with sufficient instructions to faithfully reproduce the main experimental results, as described in supplemental material?
    \item[] Answer: \answerYes{}
    \item[] Justification: We provide anonymized public repositories for the dataset and code, including the benchmark files, candidate graph files, Croissant metadata, README documentation, dependencies, and commands to reproduce the reported evaluations.

\item {\bf Experimental setting/details}
    \item[] Question: Does the paper specify all the training and test details (e.g., data splits, hyperparameters, how they were chosen, type of optimizer) necessary to understand the results?
    \item[] Answer: \answerYes{}
    \item[] Justification: Section~\ref{sec:bench:eval} specifies the evaluation-only setting, model and baseline configurations, decoding parameters, candidate-path construction, answer extraction rules, and metrics. No model training is performed in the main experiments.

\item {\bf Experiment statistical significance}
    \item[] Question: Does the paper report error bars suitably and correctly defined or other appropriate information about the statistical significance of the experiments?
    \item[] Answer: \answerNo{}
    \item[] Justification: We report deterministic aggregate accuracy and per-type breakdowns but do not include error bars, because the benchmark is fixed and most experiments use deterministic or near-deterministic decoding. We acknowledge this as a limitation and release the evaluation scripts to enable future repeated-run analyses.

\item {\bf Experiments compute resources}
    \item[] Question: For each experiment, does the paper provide sufficient information on the computer resources (type of compute workers, memory, time of execution) needed to reproduce the experiments?
    \item[] Answer: \answerYes{}
    \item[] Justification: Section~\ref{sec:resource} reports the compute environment used for dataset construction and evaluation, 
    
\item {\bf Code of ethics}
    \item[] Question: Does the research conducted in the paper conform, in every respect, with the NeurIPS Code of Ethics \url{https://neurips.cc/public/EthicsGuidelines}?
    \item[] Answer: \answerYes{}
    \item[] Justification: The research uses publicly available knowledge graph data, releases only benchmark and evaluation resources, and documents intended use, limitations, licensing, and potential risks. 

\item {\bf Broader impacts}
    \item[] Question: Does the paper discuss both potential positive societal impacts and negative societal impacts of the work performed?
    \item[] Answer: \answerYes{}
    \item[] Justification: Section~\ref{sec:limitation} discusses positive impacts such as more realistic KGQA evaluation and negative impacts such as reinforcing Wikidata biases or over-interpreting benchmark accuracy as real-world reasoning competence.
    
\item {\bf Safeguards}
    \item[] Question: Does the paper describe safeguards that have been put in place for responsible release of data or models that have a high risk for misuse (e.g., pre-trained language models, image generators, or scraped datasets)?
    \item[] Answer: \answerNA{}
    \item[] Justification: The paper does not release high-risk models, scraped images, or dual-use generative systems. The released benchmark is documented with intended use, limitations, license information, and Responsible AI metadata.

\item {\bf Licenses for existing assets}
    \item[] Question: Are the creators or original owners of assets (e.g., code, data, models), used in the paper, properly credited and are the license and terms of use explicitly mentioned and properly respected?
    \item[] Answer: \answerYes{}
    \item[] Justification: We cite Wikidata and the baseline methods used in the evaluation, and document the licenses and terms of the released benchmark.

\item {\bf New assets}
    \item[] Question: Are new assets introduced in the paper well documented and is the documentation provided alongside the assets?
    \item[] Answer: \answerYes{}
    \item[] Justification:  The paper introduces and releases a new benchmark and evaluation code. The dataset card, Croissant metadata with Responsible AI fields, README files, schema descriptions, license, and usage instructions are provided alongside the assets.

\item {\bf Crowdsourcing and research with human subjects}
    \item[] Question: For crowdsourcing experiments and research with human subjects, does the paper include the full text of instructions given to participants and screenshots, if applicable, as well as details about compensation (if any)? 
    \item[] Answer: \answerNA{}
    \item[] Justification: The paper does not involve crowdsourcing, external annotators, user studies, or experiments with human participants. Author-side dataset verification was conducted as part of benchmark construction and does not constitute human-subject research.

\item {\bf Institutional review board (IRB) approvals or equivalent for research with human subjects}
    \item[] Question: Does the paper describe potential risks incurred by study participants, whether such risks were disclosed to the subjects, and whether Institutional Review Board (IRB) approvals (or an equivalent approval/review based on the requirements of your country or institution) were obtained?
    \item[] Answer: \answerNA{}
    \item[] Justification: The study does not involve crowdsourcing or human-subject experiments, so IRB approval or equivalent review is not applicable.

\item {\bf Declaration of LLM usage}
    \item[] Question: Does the paper describe the usage of LLMs if it is an important, original, or non-standard component of the core methods in this research? Note that if the LLM is used only for writing, editing, or formatting purposes and does \emph{not} impact the core methodology, scientific rigor, or originality of the research, declaration is not required.
    \item[] Answer: \answerYes{}
    \item[] Justification: Sections~\ref{sec:method:pipeline} describe how LLMs are used for question generation, filtering, answer selection, closed-book baselines, and pseudoword analysis. We report the model names, prompts or prompt templates, and decoding settings used in the experiments.

\end{enumerate}

\end{document}